\documentclass[conference]{IEEEtran}
\IEEEoverridecommandlockouts

\usepackage{cite}
\usepackage{amsmath,amssymb,amsfonts}
\usepackage{algorithmic}
\usepackage{graphicx}
\usepackage{textcomp}
\usepackage{xcolor}
\usepackage{multirow}
\usepackage{threeparttable}
\usepackage{booktabs}
\usepackage{epstopdf}
\usepackage[T1]{fontenc}
\usepackage{aecompl}
\usepackage{cleveref}
\usepackage{tabularx}
\usepackage{nicematrix,tikz}
\usepackage{color, soul}

\def\BibTeX{{\rm B\kern-.05em{\sc i\kern-.025em b}\kern-.08em
    T\kern-.1667em\lower.7ex\hbox{E}\kern-.125emX}}

\begin{document}

\title{Transfer Learning-based State of Health Estimation for Lithium-ion Battery with Cycle Synchronization\\
}

\author{Kate Qi~Zhou,~\IEEEmembership{}
	Yan~Qin,~\IEEEmembership{Member,~IEEE,}
	Chau~Yuen,~\IEEEmembership{Fellow,~IEEE}
	\thanks{This work was supported by EMA-EP011-SLEP-001. (Corresponding author: Yan Qin)}
	\thanks{K. Q. Zhou, Y. Qin, and C. Yuen are with the Engineering Product Development Pillar, The Singapore University of Technology and Design, 8 Somapah Road, 487372 Singapore. (e-mail: qi$\_$zhou@mymail.sutd.edu.sg, yan.qin@ntu.edu.sg, yuenchau@sutd.edu.sg)}}


\onecolumn
\maketitle
\begin{abstract}
Accurately estimating a battery's state of health (SOH) helps prevent battery-powered applications from failing unexpectedly. With the superiority of reducing the data requirement of model training for new batteries, transfer learning (TL) emerges as a promising machine learning approach that applies knowledge learned from a source battery, which has a large amount of data. However, the determination of whether the source battery model is reasonable and which part of information can be transferred for SOH estimation are rarely discussed, despite these being critical components of a successful TL. To address these challenges, this paper proposes an interpretable TL-based SOH estimation method by exploiting the temporal dynamic to assist transfer learning, which consists of three parts. First, with the help of dynamic time warping, the temporal data from the discharge time series are synchronized,  yielding the warping path of the cycle-synchronized time series responsible for capacity degradation over cycles. Second,  the canonical variates retrieved from the spatial path of the cycle-synchronized time series are used for distribution similarity analysis between the source and target batteries.  Third, when the distribution similarity is within the predefined threshold,  a comprehensive target SOH estimation model is constructed by transferring the common temporal dynamics from the source SOH estimation model and compensating the errors with a residual model from the target battery.  Through a widely-used open-source benchmark dataset, the estimation error of the proposed method evaluated by the root mean squared error is as low as 0.0034 resulting in a 77$\%$ accuracy improvement compared with existing methods.

\end{abstract}


\section{Introduction}
Recyclable lithium-ion battery (LiB), as the primary energy storage component used in electronics products like mobile phones, computers, and electric cars \cite{58}, serves a vitally critical role in reducing the global carbon footprint and combating the climate crisis \cite{61}.  It has the characteristics of high energy density,  and outstanding high-temperature performance \cite{57}. The state of health (SOH) of LiB, commonly indicated by its capacity,  degrades during charging and discharging \cite{33} \cite{53}, reflected by the capacity loss or the resistance increment. For device performance management and maintenance, the ability to accurately estimate the LiB's SOH is critical.

Data-driven methods based on machine learning have been widely applied to estimate the SOH during battery degradation. It has the benefit of avoiding the examination of the internal battery mechanics like the model-based technique \cite{56} \cite{34}. It utilizes the information extracted from recorded or real-time measurement \cite{55} and estimates the regression connection by mapping nonlinear functions \cite{51}. Support Vector Machine (SVM) \cite{25}, Gaussian Process Regression (GPR) \cite{60} and, Recurrent Neural Network (RNN) \cite{48},  are most popular algorithms employed for the regression models in the literature. Specifically, RNN models represented by the long short-term memory (LSTM) \cite{3} and the gated recurrent unit (GRU) \cite{4} are popular in SOH and RUL prediction as they are good at processing long sequence data \cite{68}. For instance, Jayasinghe $et\;al.$ \cite{15} combined temporal convolution layers and LSTM layers to predict the RUL by learning the prominent characteristics and complicated temporal variations in sensor values. Zhang $et\;al. $ \cite{16} employed LSTM with Monte Carlo simulation to generate a probabilistic RUL prediction by detaining the underlying long-term dependencies during the degradation. Cui $et\;al. $ \cite{14} proposed a dynamic spatial-temporal attention-based gated recurrent unit model, and its results outperformed SVM and GPR in SOH estimation of LiB.

To make the data-driven methods work, massive data needs to be collected to train the model, which is time-consuming and expensive \cite{47}. Furthermore, due to the divergent electrochemistry reactions throughout different deterioration stages, the data collected in the past on a single battery has different data distribution from the estimation period.  Thus, the model built by the historical information might not be applicable to estimate the online SOH status.  There is a necessity for using less data yet acquiring the same information to make the estimation work \cite{52}. Transfer learning (TL) has recently come up as an emerging machine learning method to fulfill this.  It utilizes the model built by the existing source battery, in which abundant data is available to cover a comprehensive system process and transfer the generic features to the new prediction task on the target battery \cite{59}. Applying TL to battery SOH estimation saves the cost of collecting data on the target battery, shortens the time to retrain the model from scratch, and improves the accuracy at the same time.  Shu $et\;al. $ \cite{7} integrated TL with a fine-tuning strategy to predict SOH by using the charging duration at a predefined voltage range as the health feature and proposed an improved LSTM network.  Tan $et\;al. $ \cite{8} developed a feature expression scoring rule and implemented TL to fine-tune or rebuild the last two fully connected layers based on the feature expression scoring for SOH prediction. Che $et\;al. $ \cite{9} combined TL and gated RNN, including LSTM and GRU, to train the model with the most relevant battery and fine-tune it using early cycling data of the target battery.  Kim $et\;al. $ \cite{64} proposed a VarLSTM-TL to facilitate accurate SOH forecasting and RUL prediction for different LIB types to predict RUL values at a single point in time and forecast capacity-degradation patterns with credible intervals.  Ma $et\;al. $ \cite{65} introduced a transfer learning-based method by combining a convolutional neural network (CNN) with an improved domain adaptation method that is used to construct an SOH estimation model.

Although approaches discussed above have yielded good results in battery SOH prediction,  gaps are observed ranging from feature extraction, model interpretability, and how to achieve TL, which are specified as follows:

\begin{itemize}
\item Current research uses several ways to prepare the data to meet the standard input requirement of RNN,  like selecting health indicators \cite{17},  manual truncation \cite{19},  zero-padding \cite{20},  etc.  Rough processing of battery data likely results in information loss and negatively influences the transferring performance owing to the failure to pick a portion of the data.


\item The transfer performance may be poor if the similarity between the source and target batteries is far away. By implicitly assuming the target battery is related to the source battery,  the "when to transfer" is not explicitly articulated. 


\item The current TL applications update the last one or two layers of the network by using a small set of data from the target battery.  It leaves most of the work for end-to-end models to perform feature extraction from the source battery.  In this way,  the interpretability is unclear, as the "what to transfer" information fades once the weights are updated through retraining.
\end{itemize}

To address the challenges mentioned above, this paper presents a novel TL-based SOH estimation model. First, to acquire the unified data structure, a cycle synchronization method using dynamic time warping (DTW) is proposed in this paper to transform the time index to be on the vertical axis and synchronize the uneven length of the discharge cycles. Next, to determine "when to transfer", the hidden temporal mechanism during the battery discharging process is exploited by canonical variate analysis (CVA) \cite{44}.  Statistics control limits measure the degradation distribution similarity between target and source batteries to determine the transfer capability of the source battery.  Finally, to clearly identify "what to transfer", the source SOH estimation model is constructed by using canonical variates (CVs) from the source battery,  and the target SOH estimation model is constructed by transferring the common CVs from the source SOH estimation model and improved by the target battery residual model trained by the target battery-specific residual variate. The contributions of this work are summarized as the following:

\begin{itemize}
\item Cycle synchronization is meant to deal with LiB's variable-length cycle data, overcoming data loss or distortion caused by improper data processing. 
\item Battery degradation distribution similarity analysis provides a benchmark to evaluate the source battery's ability to transfer. 
\item A hybrid target SOH estimation model is proposed by transferring the common features from the source SOH estimation model and dynamically adjusting them with a small dataset of target domain-specific features to achieve a superior estimation outcome.
\end{itemize}

The remaining parts of this paper are organized as follows. Section II illustrates battery degradation behavior and how machine learning can be applied to estimate battery SOH. Section III proposes TL-based SOH estimation with cycle synchronization in detail, and Section IV discusses the experiment results. At last, this article's conclusion and future work are presented in Section V.

\section{Preliminary}

This first section describes the LiBs' degradation behavior.  Next, high-level comparisons of traditional machine learning and TL on battery SOH estimation are provided for a better understanding.

\subsection{Battery Degradation Behavior}

The SOH of the battery is commonly defined as the ratio of its maximum available capacity to its nominal capacity \cite{63}. A fresh battery is considered to be at full capacity, which means 100$\%$ SOH. Along with the charging and discharging process, the battery deterioration follows an inconsistently repeating but comparable aging pattern \cite{5}.  Fig. 1(a) illustrates an example of the battery degradation pattern from the dataset provided by the Massachusetts Institute of Technology (MIT) and Stanford University \cite{21}. Different capacity degradation phases have different fading patterns, with the beginning being relatively stable until it reaches a point when it begins to deteriorate quickly.  For better understanding, the discharge voltage of battery CH25 is shown in Fig. 1(b) when applying a constant discharging current. It is observed that Cycle 1 has the most extensive discharging time steps when capacity is full. The time steps of Cycle 600, Cycle 850, and Cycle 984 are getting smaller due to degradation. It is worth exploring the important role of the discharging time step, which is strongly associated with the battery's SOH status. When the battery's capacity loss exceeds 20$\%$ or 30$\%$ of its nominal capacity, it is regarded to be the end of life \cite{12}.

\begin{figure}[!htb]
\centering
\includegraphics[scale=0.4]{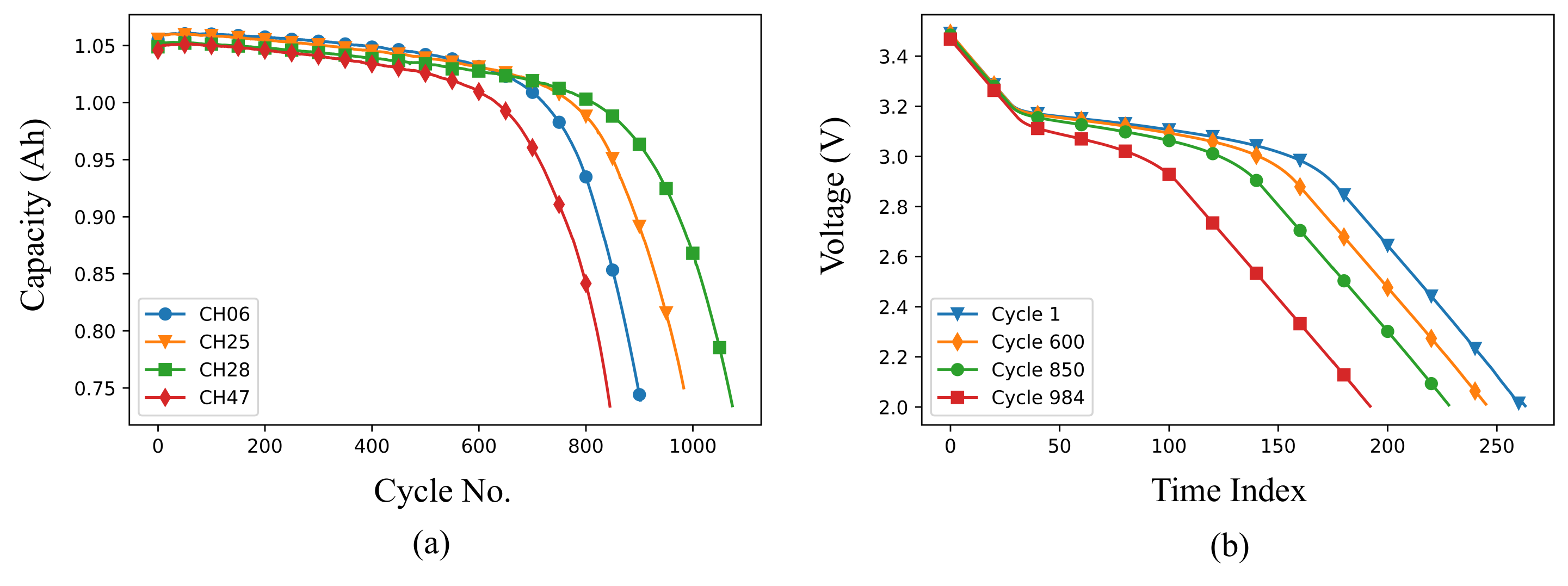}
\caption{Visualization of (a) Capacity degradation curves of four batteries, and (b) Discharge voltage samples in four selected cycles of Battery CH25.}
\label{MyFig1}
\end{figure}




\subsection{Machine Learning on Battery SOH Estimation}
Traditional machine learning uses a model trained on historical data called the source domain to predict future data, referred to as the target domain. To produce the best results, the target domain is expected to have the same data distribution as the source domain. For LiB SOH estimation, Fig. 2(a) depicts the model in which both the training and testing data are typically from the same battery, and hence they are presumed to be similar. Fig. 2(b) illustrates how the model trained on one battery is immediately applied to another battery under the premise that the data distributions of the two batteries are equal. As mentioned in Section II.A, battery degradation occurs over time and differs from battery to battery. In both circumstances, the estimation accuracy is not ideal.

TL inherits the knowledge acquired from one source learning task to the target task to enhance accuracy and efficiency, as shown in Fig. 2(c). The data distribution of source and target domains can be similar or different, but must be highly related \cite{38}. The problems of "what to transfer" and "how to transfer" are addressed by identifying the common features shared by the source and target batteries, and building a model to transfer them. To decide "when to transfer," in battery SOH estimation, it is critical to consider whether the source and target battery data distributions are highly related. It can be done by statistical approaches to extract the temporal correlation as the degradation is significantly tied to time.

\begin{figure}[htb]
\centering
\includegraphics[scale=0.25]{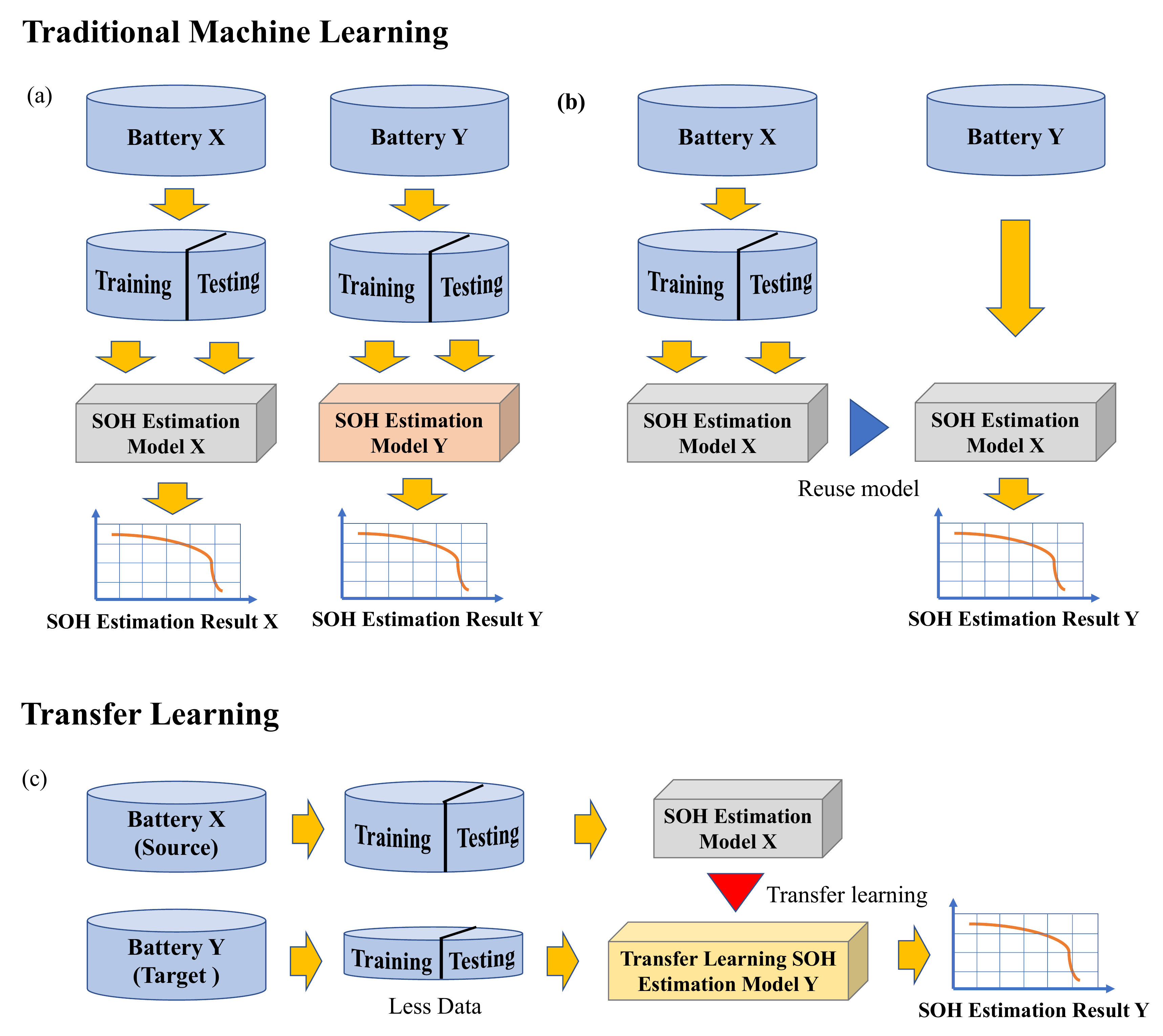}
\caption{The comparisons of battery SOH estimation between (a) traditional machine learning using its own data, (b) traditional machine learning using models trained by other batteries, and (c) transfer learning models.}
\label{MyFig2}
\end{figure}
\addtolength{\textfloatsep}{-0.15in}

\begin{figure*}[htb]
\centering
\includegraphics[scale=0.35]{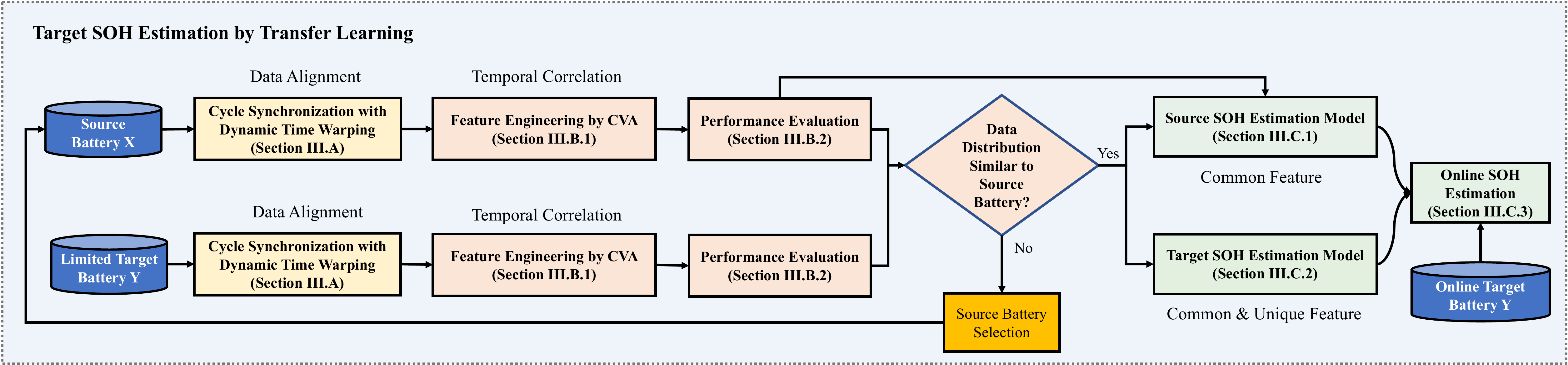}
\caption{The overall framework of the proposed method.}
\label{MyFig3}
\end{figure*}

\section{Transfer Learning-based State of Health Estimation with Cycle Synchronization}
The proposed target SOH estimation process flow is presented in Fig. 3. The entire procedure begins with cycle synchronization to convert discharge voltage cycles to the same length. Subsequently, source and target batteries' data distribution similarities are assessed. Finally, a TL-based target SOH estimation model is created by integrating the estimation from the source SOH estimation model by common features and the one from the target battery residual model by target battery unique features.

\subsection{Cycle Synchronization with Dynamic Time Warping}
During the discharging process, the discharge cycle time reduces due to battery degradation.  The discharge cycle's inconsistency is incompatible with CVA processing and RNN input. It is crucial to extract time-domain features for a continuous dynamical process \cite{50} and synchronize the cycles with their temporal information retained.  In order to achieve these purposes,  we develop cycle synchronization techniques using DTW \cite{1} to convert varying length voltage-based discharge time series into time-index-based time series.

DTW is renowned for determining the best alignment of two series by expanding and contracting localized portions to obtain the smallest possible distance between them  \cite{2}. Given a reference discharge voltage series $\mathbf v_{s}=[v_{s}^1, v_{s}^2,... ,v_{s}^i,... ,v_{s}^m]$ from a source battery as illustrated in Fig. 4(a),  a warping path $\mathbf W$ based on minimum-distance between $\mathbf v_{s}$ and the target discharge voltage series $\mathbf v_{t}=[v_{t}^1, v_{t}^2,... ,v_{t}^j,... ,v_{t}^n]$ is built by DTW, where the reference series has $m$ samples and target series has $n$ samples.  By using each sample in the time series, warping path $\mathbf W$ starts from $\mathbf w_1=(0,0)$ and ends with $\mathbf w_G=(m,  n)$. Multiple steps in the reference series can be matched to a step on the target series, and vice versa. The warping path W is denoted as,
\begin{equation*}
\begin{aligned}
\mathbf W=[w_1, w_2, ..., w_g,...,w_G]     
\end{aligned}
\end{equation*}
where $G$ is the length of the warping path, and $\max(m,n) \le G \le m+n$.

The $g^{th}$ point of the warping path where the $j^{th}$ step of the target series is mapped to the $i^{th}$ step in the reference series is,
\begin{equation*}
\begin{aligned}
w_{g}=(g_{s}^i,g_{t}^j)    
\end{aligned}
\end{equation*}
where $g_{s}^i$ is the time index on the source series and $g_{t}^j$ is the time index on the target series.

\begin{figure}[!htb]
\centering
\includegraphics[scale=0.65]{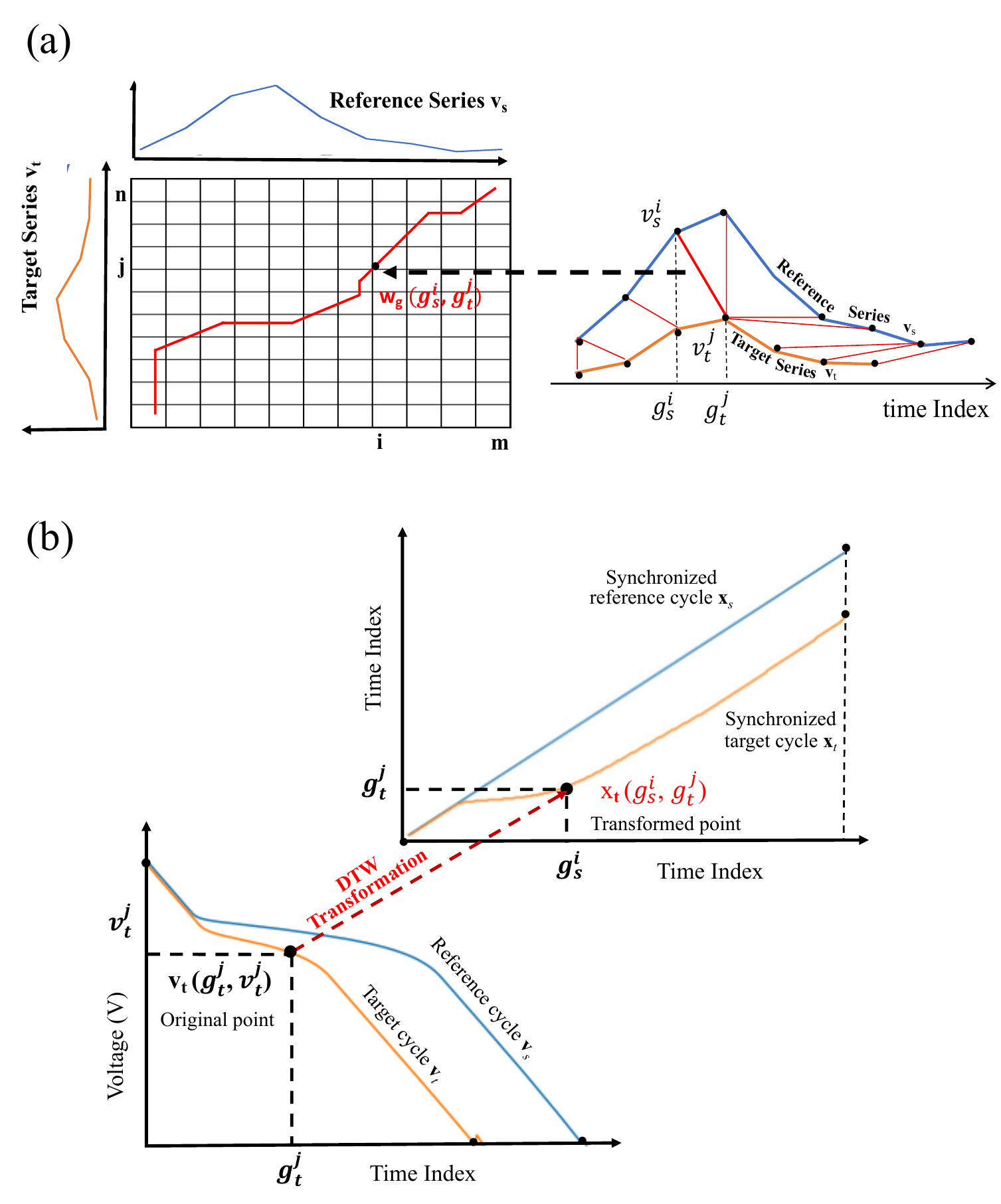}
\caption{Illustration of (a) DTW warping path,  (b) Battery discharge cycle synchronization, $v_t(g_t^j,v_t^j)$ denotes the point on the original discharge voltage cycle and $x_t(g_s^i,g_t^j)$ denotes the point on the synchronized cycle after the DTW transformation.}
\label{MyFig4}
\end{figure}

The mapping information transforms point $v_t(g_{t}^j,v_{t}^j)$, the voltage coordinate at point $j$ on the target discharge voltage cycle, to $x_t(g_{s}^i,g_{t}^j)$ on the synchronized cycle.   By putting the $g_{s}^i$ on the x-axis and the $g_{t}^j$ on the y-axis, a new series is produced.  The reference series is transformed into a diagonal line as $v_s(g_{s}^i,v_{s}^i)$ is converted into $x_s(g_{s}^i,g_{s}^i)$ putting both $g_{s}^i$ on the x-axis and y-axis, as illustrated in Fig. 4(b). For those many time steps on the target series that are mapped to a single time step on the reference series, the mean of them ${\overline g}_{t}^j$ is used. The cycle synchronized voltage time-index-based time series is created as $\mathbf x_{t}=[g_{t}^1, g_{t}^2,..., {\overline g}_{t}^j, ..., g_{t}^n] $. Finally, all the $K$ discharge voltage cycles in the source battery are converted to $\mathbf X_{s}=[\mathbf x_{s,1}, \mathbf x_{s,2},...,\mathbf x_{s,K}] $ and all the $L$ discharge voltage cycles in the target battery are converted to $\mathbf X_{t}=[\mathbf x_{t,1}, \mathbf x_{t,2},...\mathbf x_{t,L}] $. 

\setlength{\abovedisplayskip}{3pt}
\setlength{\belowdisplayskip}{3pt}

\subsection{Degradation Distribution Similarity Analysis by CVA}
\subsubsection{Feature Engineering by CVA}
CVA is a dissimilarity measurement approach that aims to maximize the correlation of two different sets of variables in terms of their within-groups variation \cite{28}. It is performed by selecting canonical variates (CVs), which are the linear combinations of one variable set that are better correlated with the linear combinations of another variable set \cite{45} \cite{62}. CVA uses the complete degradation source battery voltage synchronized cycles to generate the transformation matrices and the threshold. The transformation matrices aid in generating the CVs from the source battery to feed into the estimation model, as well as converting the target battery's high-dimensional data into an indicator, which is then compared with the threshold \cite{41} to determine how similar the source and target batteries' discharge data distributions are.

In order to consider the time correlation within a cycle,  at each sample $i$ in $\mathbf x_{s,k}$, the $k^{th}$ cycle of source battery $\mathbf X_{s}$, will be expanded by considering the past $p$ and future $f$ samples to generate the past and future vectors as $\mathbf x_{p,k}(i)$ and $\mathbf x_{f,k}(i)$,
\begin{equation*}
\mathbf x_{p,k}(i)= 
\begin{bmatrix}
x_{s,k} (i-1) \\
x_{s,k} (i-2) \\
\mathbf \vdots \\
x_{s,k} (i-p)
\end{bmatrix}
\end{equation*}

\begin{equation}
\mathbf x_{f,k}(i)= 
\begin{bmatrix}
x_{s,k} (i) \\
x_{s,k} (i+1) \\
\mathbf \vdots \\
x_{s,k} (i+f-1)
\end{bmatrix}
\end{equation}
where $p$ and $f$ are past and future lags, which is normally chosen to encapsulate the data autocorrelation \cite{46}.

The past and future vectors from each point in the cycle are put together to form the past and future Hankel matrix, namely $\mathbf X_{p,k}$ and $\mathbf X_{f,k}$,
\begin{equation}
\begin{aligned}
&\mathbf X_{p,k}=[\mathbf x_{p,k}(m-f), ... , \mathbf x_{p,k}(p+1),\mathbf x_{p,k}(p)]\\
&\mathbf X_{f,k}=[\mathbf x_{f,k}(m-f+1), ... , \mathbf x_{f,k}(p+2),\mathbf x_{f,k}(p+1)]
\end{aligned}
\end{equation}
where $m$ is the total samples in the cycle.

Finally, the complete past and future matrices $\mathbf X_{p}$ and $\mathbf X_{f}$  are created by putting $\mathbf X_{p,k}$ and $\mathbf X_{f,k}$ from all the cycles, respectively,
\begin{equation}
\begin{aligned}
&\mathbf X_{p}=[\mathbf X_{p,1},  \mathbf X_{p,2},..., \mathbf X_{p,K}]\\
&\mathbf X_{f}=[\mathbf X_{f,1},  \mathbf X_{f,2},..., \mathbf X_{f,K}]
\end{aligned}
\end{equation}
where $K$ is the total cycles of the source battery.

The same symbol $\mathbf X_{p}$ and $\mathbf X_{f}$ are still used for normalized matrices, respectively.  With the normalized $\mathbf X_{p}$ and $\mathbf X_{f}$, the covariance and cross-covariance matrices of the past and future observations can be found as the following,
\begin{equation}
\begin{aligned}
&\mathbf \Sigma_{pp}=\frac{1}{H-1}\mathbf X_{p} \mathbf X_{p}^{\rm T}\\
&\mathbf \Sigma_{ff}=\frac{1}{H-1}\mathbf X_{f} \mathbf X_{f}^{\rm T}\\
&\mathbf \Sigma_{fp}=\frac{1}{H-1}\mathbf X_{f} \mathbf X_{p}^{\rm T}
\end{aligned}
\end{equation}
where $H =K(m-f-p+1)$.

Singular value decomposition is performed accordingly to find the CVs,
\begin{equation}
\begin{aligned}
\mathbf \Sigma_{pp}^{-1/2}\mathbf\Sigma_{fp} \mathbf \Sigma_{ff}^{-1/2}=\mathbf U\mathbf\Sigma\mathbf V
\end{aligned}
\end{equation}
where $\mathbf\Sigma =diag[\alpha_1,\alpha_2,...,\alpha_s]$ is a diagonal matrix of nonnegative singular values with the order of $1 > \alpha_1>\alpha_2>,...,>\alpha_s$ , $\mathbf U$ is right-singular and $\mathbf V$ is left-singular matrix. 

The first $C$ largest $\alpha$ has exhibited the most correlated canonical variates (CV), indicating a high temporal dynamic correlation. To indicate the space to be retained, the first $C$ columns of $\mathbf V$ are chosen accordingly as $\mathbf V_c$. The remainder of the $\alpha$, which has lower values and is linked to low correlation, is grouped as residual variates (RVs) under residual space. The source transform matrix $\mathbf J_{s,c}$ and $\mathbf J_{s,r}$ to convert the past matrix to CVs and RVs are calculated as follows,

\begin{equation}
\begin{aligned}
\mathbf J_{s,c} &=\mathbf V_{c}^{\rm T} \Sigma_{pp}^{-1/2} \\
\mathbf J_{s,r} &=(\mathbf I - \mathbf V_{c}\mathbf V_{c}^{\rm T} )\Sigma_{pp}^{-1/2}
\end{aligned}
\end{equation}

By source transform matrix $\mathbf J_{s,c}$ and $\mathbf J_{s,r}$,  the source battery CV $\mathbf Z_{s,c}$ and RV $\mathbf Z_{s,r}$ are obtained as, 

\begin{equation}
\begin{aligned}
\mathbf Z_{s,c} &=\mathbf J_{s,c} \mathbf X_{p} =\mathbf V_{c}^{\rm T} \Sigma_{pp}^{-1/2} \mathbf X_{p}\\
\mathbf Z_{s,r} &=\mathbf J_{s,r} \mathbf X_{p}=(\mathbf I - \mathbf V_{c}\mathbf V_{c}^{\rm T} )\Sigma_{pp}^{-1/2}\mathbf X_{p}
\end{aligned}
\end{equation}

Source battery $\mathbf Z_{s,c}$ span the retained space while $\mathbf Z_{s,r}$ span the residue space. To quantify the variance of $\mathbf Z_{s,c}$ and $\mathbf Z_{s,r}$, the statistical indicators Hotelling $\mathbf T^2$ statistic and Q statistic are introduced,
\begin{equation}
\begin{aligned}
\mathbf T^2_s{(k)}=\sum_{i=1}^{C} \mathbf z^2_{s,c,i}(k)\\
\mathbf Q_s(k)=\sum_{i=1}^{p} \mathbf z^2_{s,r,i}(k)
\end{aligned}
\end{equation}
where $\mathbf z_{s,c,i}(k)$ and $\mathbf z_{s, r,i}(k)$ are the $k^{th}$ row and $i^{th}$ column of $\mathbf Z_{s,c}$ and $\mathbf Z_{s,r}$, respectively.

The control limits $\mathbf {CL}_{T_s^2}$ and $\mathbf {CL}_{Q_s}$ of $\mathbf T^2$ and $\mathbf Q$ are derived based on a significant level $\beta$ to cover a certain percentage of data. In this paper, the probability distribution is estimated using kernel density estimations,  and its control limits are calculated accordingly\cite{39}.

\subsubsection{Similarity Evaluation}
Due to internal electrochemical reactions during charging and discharging, LiB deterioration varies vastly. The information from the source battery may not be suitable to be applied to the target battery if their degradations are dissimilar.  The degradation distribution variability in the source retained space and residual space within each cycle is reflected by $\mathbf T_s^2$, which represents the total variation in the retained space, and $\mathbf Q_s$, which represents the sum of the squared variation errors in the residual space. Therefore,  the control limits $\mathbf {CL}_{T_s^2}$ and $\mathbf {CL}_{Q_s}$ generated from $T_s^2$ and $Q_s$ are employed as significant level indicators.  


As a result,  an evaluation method based on canonical $\mathbf {CL}_{T_s^2}$ and residual $\mathbf {CL}_{Q_s}$ is presented to benchmark the similarity between the target and source batteries and choose the appropriate estimation model accordingly. The steps are as follows:

Step 1: Following the methods in Section III.A, the target battery's limited discharge cycle data is converted to a time-index series under the same reference cycle as the source battery.

Step 2: Using the same time lags $p$ and $f$, past matrix $\mathbf X_{t,p}$ are generated through Eqs. (1) to (3).  To convert the target battery's discharge voltage data to the same baseline as the source battery's, the CVs $\mathbf Z_{t,c}$ and RVs $\mathbf Z_{t,r}$ are obtained by projecting $\mathbf X_{t,p}$ onto the source transform matrix $\mathbf J_{s,c}$ and residual matrix $\mathbf J_{s,r}$,
\begin{equation}
\begin{aligned}
\mathbf Z_{t,c} &=\mathbf J_{s,c} \mathbf X_{t,p}\\
\mathbf Z_{t,r} &=\mathbf J_{s,r} \mathbf X_{t,p}
\end{aligned}
\end{equation}

\begin{figure}[htb]
\centering
\includegraphics[scale=0.35]{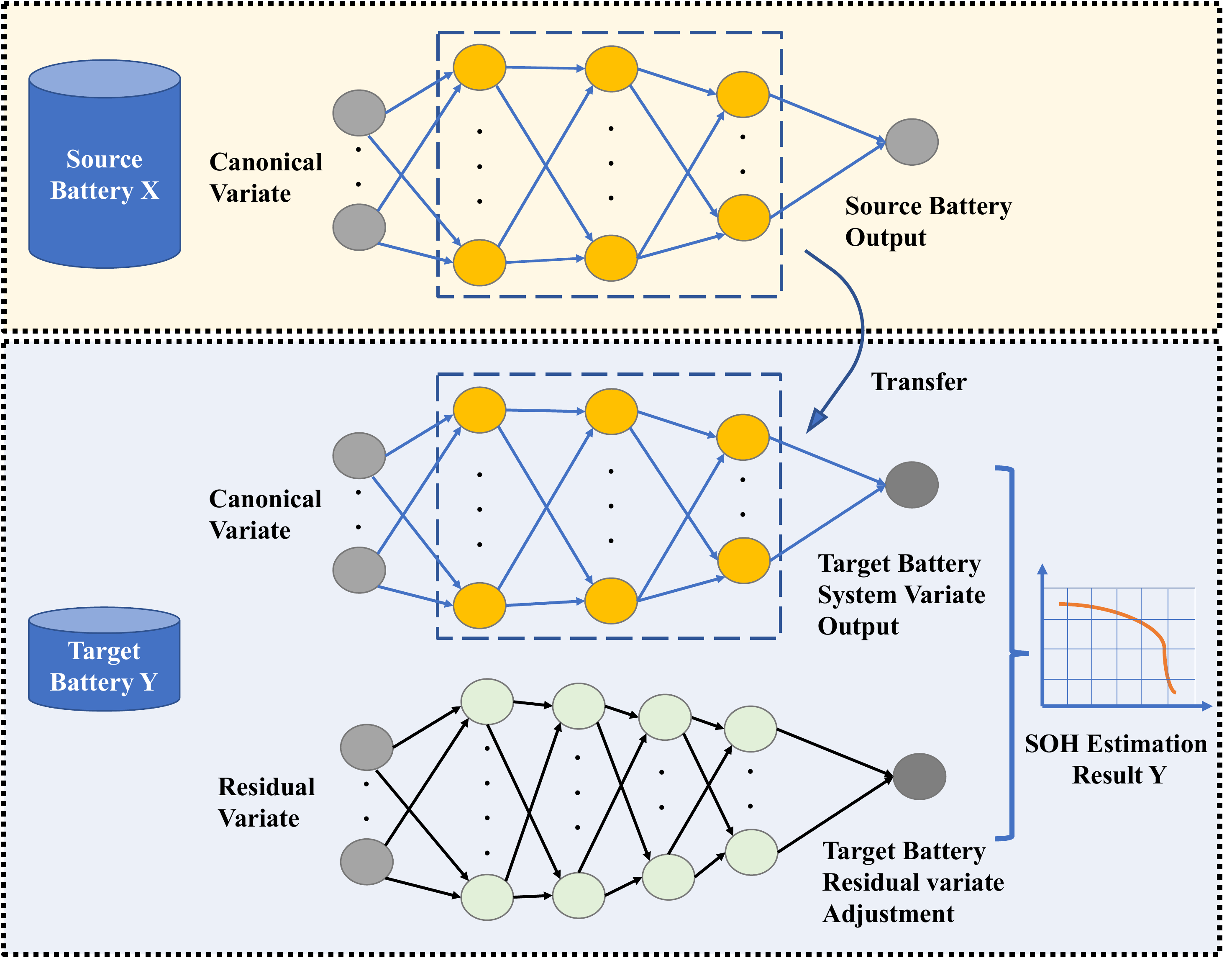}
\caption{Illustration of the development of target battery SOH estimation model}
\label{MyFig5}
\end{figure}

Step 3: Using the same number of singular values $C$ derived from the source battery, the statistical indicators $\mathbf T_t^2$ and $\mathbf Q_t$ are calculated as Eq. (10). Simultaneously, their control limit $\mathbf {CL}_{T^2_t}$ and $\mathbf {CL}_{Q_t}$ are developed under the same significant level $\beta$ as the source battery.

\begin{equation}
\begin{aligned}
\mathbf T^2_t{(k)}=\sum_{i=1}^{C} \mathbf z^2_{t,c,i}(k)\\
\mathbf Q_t(k)=\sum_{i=1}^{p} \mathbf z^2_{t, r,i}(k)
\end{aligned}
\end{equation}

Step 4: Being converted on the same basis as the source battery,  the control limit of the target battery data can be used to compare the corresponding cycles to determine the similarity with the source battery.   For the first 100 discharge voltage cycles, control limits $\mathbf {CL}_{T^2}$ and $\mathbf {CL}_{Q}$ are separated as Scenario 1 (S1) and Scenario 2 (S2). S1 examines if the control limit $\mathbf {CL}_{T^2_t}$ is within the 15$\%$ error zone of the source battery control limits $\mathbf {CL}_{T^2_s}$ for 90$\%$ of the cycles,  while S2 compares $\mathbf {CL}_{Q_t}$ with $\mathbf {CL}_{Q_s}$ under the same guideline. When both S1 and S2 are met, the target battery's degradation pattern is considered to be similar to that of the source battery. 

\subsection{Target SOH Estimation by Transfer Learning}
Two critical factors in TL, "what to transfer" and "how to transfer", are discussed in this section in order to achieve an outstanding estimate result.  


\subsubsection{Source SOH Estimation Model}
The source SOH estimation model $\mathbf F_s(\cdot)$ is developed using source battery $\mathbf Z_{s,c}$ from Eq. (7) as input.  

The estimation model contains one input layer, multilayers of the GRU network, and an output layer to estimate the battery capacity. The model is trained by minimizing the loss function between the estimation and the actual value,

\begin{equation}
\begin{aligned}
&\min\sum{(Y_{s}(k)-\hat{Y}_{s}(k))}^2\\
\end{aligned}
\end{equation}
\begin{equation}
\begin{aligned}
&\mathbf {s.t. } \hat{Y}_{s}(k)=\mathbf F_s(\mathbf z_{s,c}(k))
\end{aligned}
\end{equation}
where $\mathbf z_{s,c}(k)$ is the $k^{th}$ row of $\mathbf Z_{s,c}$,  $\hat{Y}_{s}(k)$ is the estimated capacity and $Y_{s}(k)$ is the measured capacity. 

The source SOH estimation model is trained by the complete life cycle data; it delivers comprehensive degradation information to the target battery to serve as the base model for target battery SOH estimation. However, due to the specific degradation path unique to each target battery,  the model performance on each target battery requires further adjustment.

\subsubsection{Target SOH Estimation Model}
By feeding a small amount of CVs $\mathbf Z_{t,c}$ from the target battery into the source SOH estimation model Eq. (12), the output $\mathbf Y_{t,c(k)}$ estimates the capacity affected by the system CVs. To further improve the estimation accuracy,  a target battery-specific residual model $\mathbf F_t(\cdot)$ is trained by feeding the RVs $\mathbf Z_{t,r}$ as input and minimizing the gap between the measured value of the capacity $Y_{t}(k)$ and the estimated capacity generated from $\mathbf F_s(\mathbf z_{t,c}(k))$,

\begin{equation}
\begin{aligned}
&\min\sum{(Y_{t}(k)-\hat{Y}_{t}(k))}^2\\
\end{aligned}
\end{equation}
\begin{equation}
\begin{aligned}
&\mathbf {s.t. } \hat{Y}_{t}(k)=\mathbf F_s(\mathbf z_{t,c}(k))+\mathbf F_t(\mathbf z_{t,r}(k))
\end{aligned}
\end{equation}
where $\mathbf z_{t,c}(k)$ is the $k^{th}$ row of $\mathbf Z_{t,c}$,  $\mathbf z_{t,r}(k)$ is the $k^{th}$ row of $\mathbf Z_{t,r}$, $Y_{t}(k)$ is the $k^{th}$ cycle of measured capacity $\mathbf Y_{t}$, and $\hat{Y}_{t}(k)$ is the estimated capacity of $Y_{t}(k)$.

The proposed method, which is a lightweight LSTM network, has a training time complexity of per time step o($W$)\cite{67},  where $W$ is the total number of weights in the network, and the model development is shown in Fig. 5.

\subsubsection{Online Target SOH Estimation}
Once the residual model $\mathbf F_t(\cdot)$ is trained by limited target battery data,  the target SOH estimation model is ready to use . When a new discharge cycle occurs, the voltage data will be used to create $\mathbf Z_{t,c,new}$ and $\mathbf Z_{t,r,new}$ by performing Steps 1 and 2 in Section III.B.  For the new cycle, the estimated capacity is achieved by Eq. (14) as below,
\begin{equation}
\begin{aligned}
 \hat{Y}_{t,new}=\mathbf F_s(\mathbf z_{t,c,new})+\mathbf F_t(\mathbf z_{t,r,new})
\end{aligned}
\end{equation}

\textit{Remark}: It is possible that a certain target battery may differ from any source batteries. In such an extreme situation, we recommend increasing the number of source batteries as many as possible. Besides, it is worth pointing out that the selected common features from the source battery will contribute to the estimation accuracy improvement, even though the target battery is different from the source battery.


\section{Experiment Result and Discussion}
The LiFePO4 cells from the dataset [30],  manufactured by A123 Systems (APR18650M1A), were put in a forced convection temperature chamber set to 30$^{\circ}$C.   They are under 10-minute fast-charging protocols with one of 224 six-step. The charging protocols have the format “CC1-CC2-CC3-CC4”.  They were discharged constantly at 4C until the voltage dropped from 3.3V to 2V. This paper uses 39 battery cells in Batch 9 under eight charge protocols, and their specifications are summarized in Table I. The battery cells with the highest starting capacity and the most extended discharge cycle from the raw data from 4 different fast-charging protocols are used as source batteries, namely CH01, CH21, CH25, and CH37. Due to the batteries being drained at a consistent current at a set temperature, only voltage data is used to estimate capacity.

\begin{table}[!htb]
	\scriptsize
	\renewcommand{\arraystretch}{1.2}
	\caption{Specification of the Batteries }
	\vspace{-2mm}
	\label{table_1}
	\begin{center}
			\begin{tabular}{p{0.23\textwidth}>{\centering\arraybackslash}m{0.2\textwidth}}
				\hline
				\toprule  
 				\multicolumn{1}{c} {\textbf {Battery Information}} & \multicolumn{1} {c}{\textbf{Detail}} \\
				\toprule  
				Battery Type & APR18650M1A \\
				Nominal Capacity & 1.1Ah \\
				Voltage Range & 2.0V \textasciitilde \ 3.6V  \\
				Discharge Current & 4C \\
				Charge Protocol 1: 4.4-5.6-5.2-4.252C & CH08, CH15, CH18, CH32 \\
				Charge Protocol 2: 3.6-6.0-5.6-4.755C & CH11,CH12,CH27,CH29,CH38 \\
				Charge Protocol 3: 4.8-5.2-5.2-4.16C & CH01,CH02,CH10,CH20,CH42 \\
				Charge Protocol 4: 5.2-5.2-4.8-4.16C & CH06,CH07,CH37,CH41,CH45 \\
				Charge Protocol 5: 6-5.6-4.4-3.834C & CH09,CH21,CH22,CH31,CH36 \\
				Charge Protocol 6: 7-4.8-4.8-3.652C & CH03,CH25,CH26,CH28,CH44 \\
				Charge Protocol 7: 8.0-4.4-4.4-3.94C & CH13,CH16,CH23,CH24,CH47 \\
				Charge Protocol 8: 8.0-7.0-5.2-2.68C & CH19,CH33,CH34,CH40,CH43 \\
				\toprule
			\end{tabular}
	\end{center}
\vspace{-4mm}
\end{table}


The root mean squared error (RMSE) is minimized to train the model. Furthermore, both RMSE and mean absolute error (MAE) are employed to measure estimate accuracy, as illustrated in the calculation below,
\begin{equation}
\begin{aligned}
\mathbf {MAE}=\frac{1}{K}\sum_{k=1}^{K}{|(Y(k) -\hat{Y}(k)|}
\end{aligned}
\end{equation}
\begin{equation}
\begin{aligned}
\mathbf {RMSE}=\sqrt{\frac{1}{K}\sum_{k=1}^{K}{(Y(k) -\hat{Y}(k))^2}}
\end{aligned}
\end{equation}
where $Y(k)$ is the measured capacity, $\hat{Y}(k)$ is the estimated capacity, and $K$ is the total number of all the discharge cycles.

\begin{figure}[htb]
\centering
\includegraphics[scale=0.4]{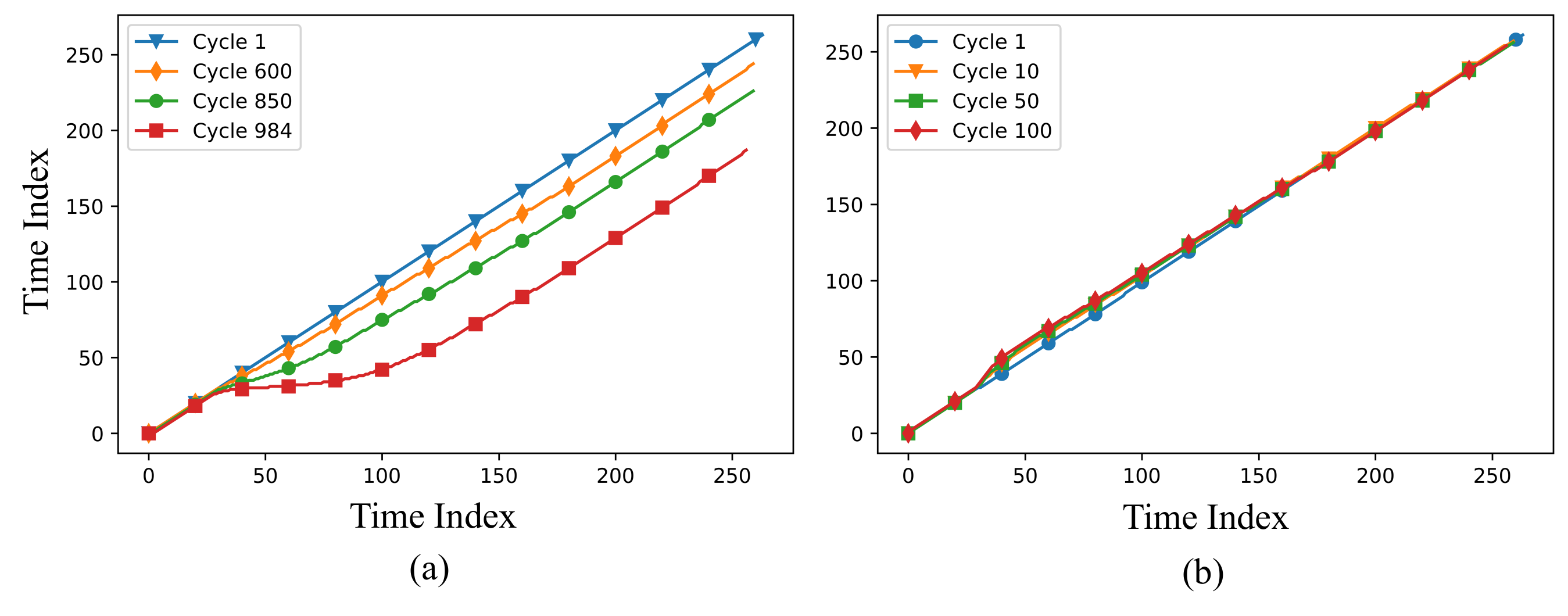}
\caption{Illustration of Time-Index based time series of (a) Source battery CH25, (b) Target battery CH28}
\label{MyFig6}
\end{figure}

\begin{figure*}[htb]
\centering
\includegraphics[scale=0.52]{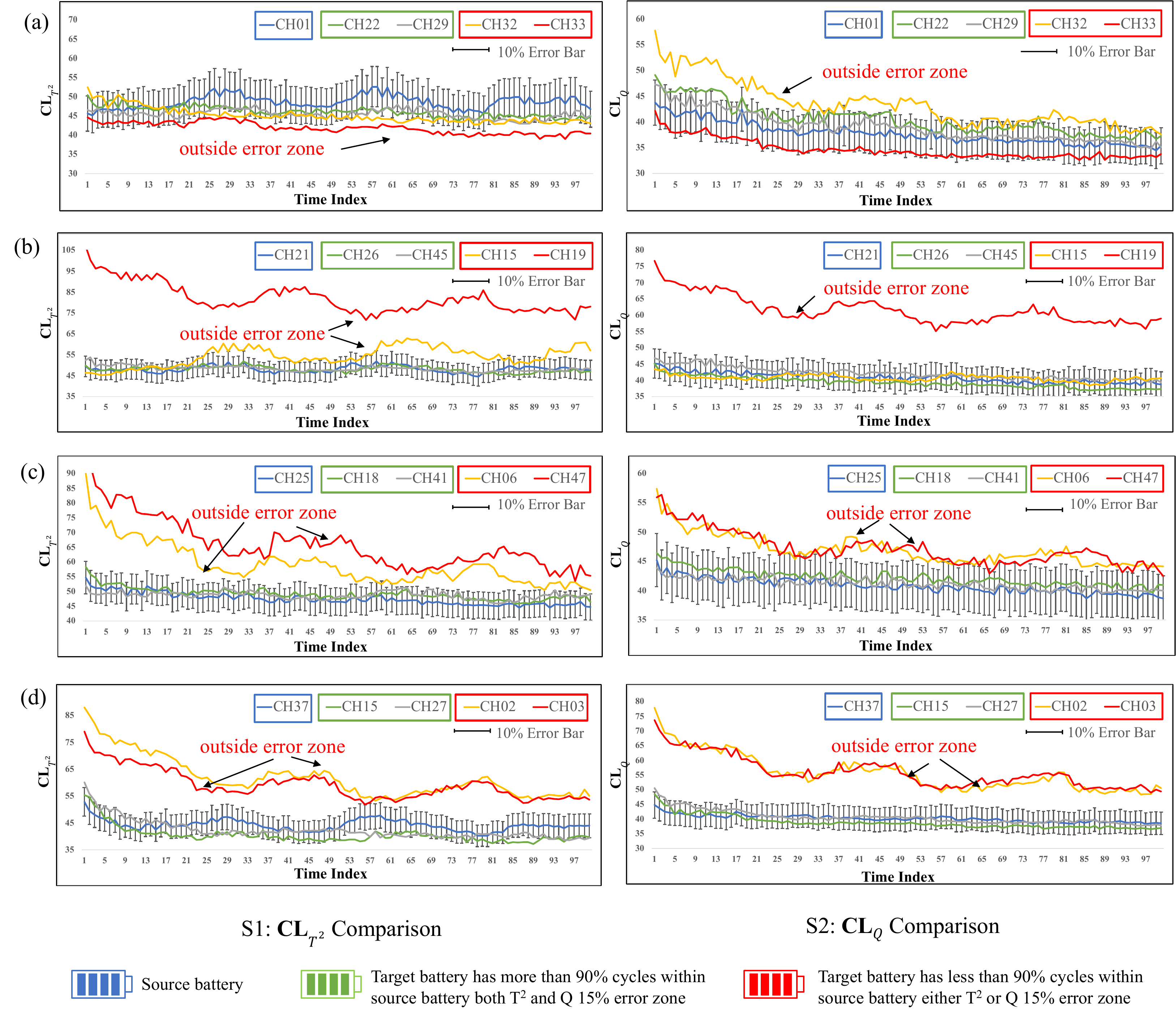}
\caption{$\mathbf {CL}_{T^2}$ and $\mathbf {CL}_{Q}$ comparison of target battery and source battery (a) Source battery CH01, (b) Source battery CH21, (c) Source battery CH25, (d) Source battery CH37.}
\label{MyFig7}
\end{figure*}

\subsection{Cycle Synchronization with Dynamic Time Warping}
Following the procedure in Section III.A,  the original time series are synchronized. The source battery's first discharge cycle is chosen as the reference cycle, during which the battery is deemed to be at full capacity.  Taking source battery CH25 for illustration, the voltage-based discharge cycles are shown in Fig. 1(b). Cycle 1 has 264 samples.  As shown in Fig. 6(a), the other discharge cycles with different time steps are translated to the time-index-based time series with the same steps of 264.  The target battery's first 100 discharge cycles are used to benchmark its similarity with the source battery.  DTW converted these target batteries' voltage-based time series to time-index-based time series based on the first discharge cycle of the source battery.  As an illustration, the results of the first 100 discharge cycles of target battery CH28 are given in Fig. 6(b), which are transformed into 264 steps as source battery CH25.  

\subsection{Degradation Distribution Similarity Analysis by CVA}
\subsubsection{Feature Extraction by CVA}

Following Eq. (1) to Eq. (7), CVs and RVs are extracted by CVA from the time-index-based time series. The autocorrelation function of the lag to have a certain impact is 25 with a 95$\%$ confidence level.  By adding a robust margin, time lags $p$ and $f$ are chosen as 32.  To determine the the number $C$ of retained CVs,  the singular values are accumulated and the curve is plotted as Fig. 8. To find the transition of the curve,  the early 15 points and last 5 points are fit using linear regression, and the connected point of the two lines are located \cite{66}.  The connected point will be the location to determine the number $C$ of retained CVs.  The four source batteries CH01, CH21, CH25, and CH37 have 20, 19, 20, and 20 retained CVs, respectively.

At the same time, control limit $\mathbf {CL}_{T^2_s}$ and $\mathbf {CL}_{Q_s}$ based on 95$\%$ significant level are obtained for the source battery each cycle. The transform matrix $\mathbf J_{s,c}$ and $\mathbf J_{s,r}$ are reserved for later target battery CVs and RVs generation.


\subsubsection{Performance Evaluation}
The performance of the source SOH estimation model is evaluated here to decide whether the transfer learning is triggered or not by comparing $\mathbf T^2$ and $\mathbf Q$ with their respective control limits $\mathbf {CL}_{T_s^2}$ and $\mathbf {CL}_{Q_s}$.  Following that, the $\mathbf Z_{t,c}$ and $\mathbf Z_{t,r}$ are extracted and $\mathbf {CL}_{T^2_t}$ and $\mathbf {CL}_{Q_t}$ are obtained for each cycle based on Eqs. (9) and (10).  For the first 100 cycles, the $\mathbf {CL}_{T^2_t}$ with $\mathbf {CL}_{T^2_s}$ under S1 and $\mathbf {CL}_{Q_t}$ with $\mathbf {CL}_{Q_s}$ under S2 are compared.  If both S1 and S2 are met, it is considered the degradation distribution of the target battery to be similar to the source battery.

\begin{figure}[htb]
\centering
\includegraphics[scale=0.4]{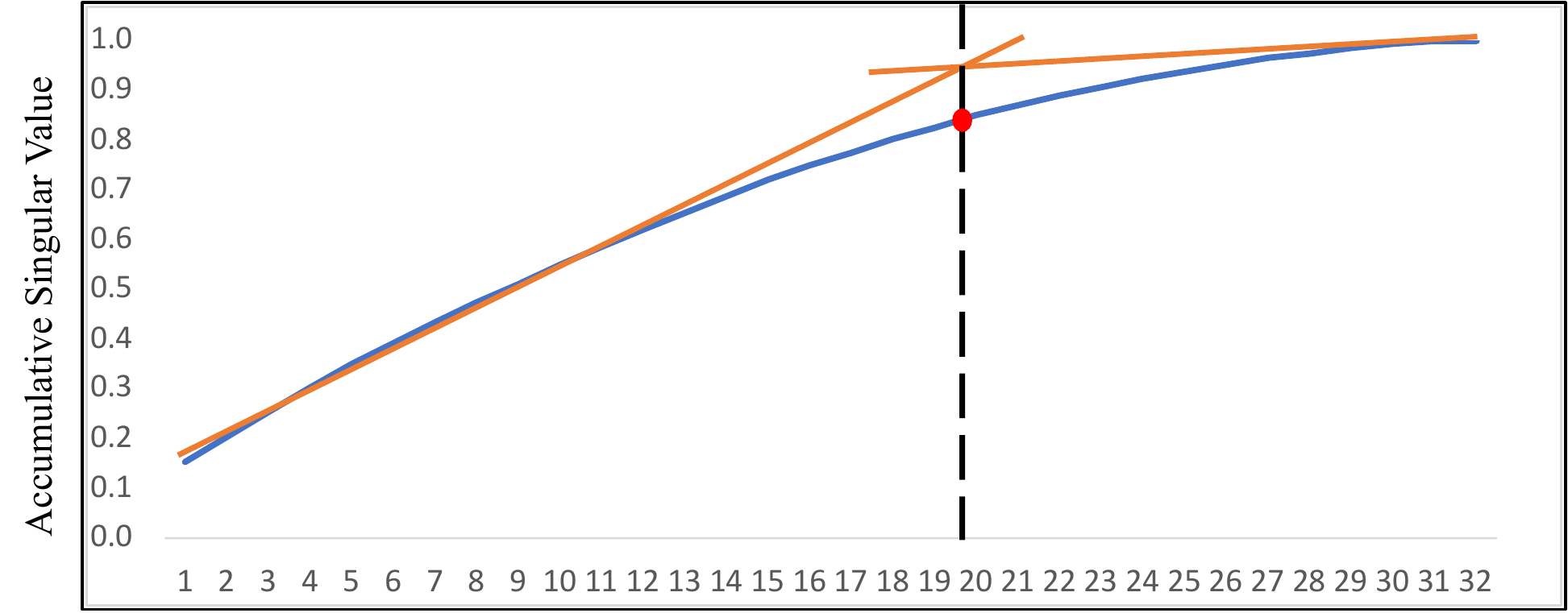}
\caption{Selection of number (C) of retained CVs for source battery CH25}
\label{MyFig8}
\end{figure}

Fig. 7 depicts the comparison results for two target batteries that are comparable to the source battery and two target batteries that are not similar, with the others omitted for brevity.  For source battery CH01, under S1, that $\mathbf {CL}_{T^2_t}$ of target battery CH33 has less than 90 cycles within $\mathbf {CL}_{T^2_s}$ 15$\%$ error zone, whereas under S2, that $\mathbf {CL}_{Q_t}$ of target battery CH32 has less than 90 cycles within $\mathbf {CL}_{Q_s}$, both of their data distribution are considered not similar as CH01, despite under the other scenario, they are within.  The same rule applies to CH15 for source battery CH21.  For target battery CH19 to source battery CH21, target battery CH06, CH47 to source battery CH25, target battery CH02, CH03 to source battery CH37, they are classified under not similar as they fail in both S1 and S2.   It is not effective to use the source SOH estimation model on those target batteries.  On the other hand,  for target battery CH22, CH29 to source battery CH01, target battery CH26, CH45 to source battery CH21,  target battery CH28, CH41 to source battery CH25, target battery CH15, CH27 to source battery CH37, they are meeting both S1 and S2.  The source SOH estimation model can serve as a good base model to transfer the knowledge learned by complete degradation cycles.  The details are shown in Table II. 

\subsection{Target SOH Estimation by Transfer Learning}
\subsubsection{Source SOH Estimation Model}
The source SOH estimation model is trained by the entire life cycle and transfers the extracted common features to the target battery,  explicitly stating "what to transfer".  The GRU-based estimation model DCVA-GRU is constructed by a GRU layer of 300 neurons and another GRU layer of 500 neurons. Before the output layer, the two GRU layers are fully linked to a normal layer with 100 neurons. Adam is chosen as the optimizer in this case.  Ninety epochs are conducted to discover the best weights and biases. 


\begin{table*}[!htb]
	\scriptsize
	\renewcommand{\arraystretch}{1.1}
	\caption{Source SOH Estimation Performance Evaluation }
	\vspace{-3mm}
	\label{table_3}
	\begin{center}
	\setlength{\tabcolsep}{3.2pt}
			\begin{tabular}{>{\centering\arraybackslash}m{0.08\textwidth}>{\centering\arraybackslash}m{0.08\textwidth}>{\centering\arraybackslash}m{0.08\textwidth}>{\centering\arraybackslash}m{0.10\textwidth}|>{\centering\arraybackslash}m{0.08\textwidth}>{\centering\arraybackslash}m{0.08\textwidth}>{\centering\arraybackslash}m{0.10\textwidth}|>{\centering\arraybackslash}m{0.08\textwidth}>{\centering\arraybackslash}m{0.08\textwidth}>{\centering\arraybackslash}m{0.08\textwidth}}
				\hline
				\toprule  
 				\multicolumn{4}{c|}{\textbf {Source Battery}} &\multicolumn{3}{c|}{\textbf{Target Battery}} &\multicolumn{3}{c}{\textbf{Evaluation}}\\
				\toprule 
				Channel& Total Cycles& Initial Capacity& Number ($C$) of retained CVs&Channel&Total Cycles&Initial Capacity&S1*&S2*&Similarity\\
				\toprule  
				\multirow{4}{2em}{\centering CH01} &\multirow{4}{2em}{\centering 958}&\multirow{4}{2em}{\centering 1.061}&\multirow{4}{2em}{\centering 20}&CH22&834&1.065&Yes&Yes&$\mathbf {Yes}$\\
				&&&&CH29&805&1.044&Yes&Yes&$\mathbf {Yes}$\\
				&&&&CH32&1074&1.045&Yes&No&No\\
				&&&&CH33&501&1.032&No&Yes&No\\
				\toprule 
				\multirow{4}{2em}{\centering CH21} &\multirow{4}{2em}{\centering 885}&\multirow{4}{2em}{\centering 1.057}&\multirow{4}{2em}{\centering 19}&CH26&917&1.039&Yes&Yes&$\mathbf {Yes}$\\
				&&&&CH45&1166&1.056&Yes&Yes&$\mathbf {Yes}$\\
				&&&&CH15&1006&1.036&Yes&No&No\\
				&&&&CH19&497&1.057&No&No&No\\
				\toprule 
				\multirow{4}{2em}{\centering CH25} &\multirow{4}{2em}{\centering 984}&\multirow{4}{2em}{\centering 1.055}&\multirow{4}{2em}{\centering 20}&CH28&1075&1.049&Yes&Yes&$\mathbf {Yes}$\\
				&&&&CH41&990&1.049&Yes&Yes&$\mathbf {Yes}$\\
				&&&&CH06&903&1.055&No&No&No\\
				&&&&CH47&762&1.048&No&No&No\\
				\toprule 
				\multirow{4}{2em}{\centering CH37} &\multirow{4}{2em}{\centering 957}&\multirow{4}{2em}{\centering 1.065}&\multirow{4}{2em}{\centering 20}&20CH15&1006&1.036&Yes&Yes&$\mathbf {Yes}$\\
				&&&&CH27&995&1.050&Yes&Yes&$\mathbf {Yes}$\\
				&&&&CH02&844&1.054&No&No&No\\
				&&&&CH03&823&1.054&No&No&No\\
				\hline
				\toprule
			\end{tabular}
	\end{center}
	\vspace{-2mm}
\begin{tablenotes}\scriptsize
\item[] *S1: 90$\%$ of $\mathbf {CL}_{T^2_t}$ Within $\mathbf {CL}_{T^2_s}$ Error Zone\\
*S2: 90$\%$ of $\mathbf {CL}_{Q_t}$ Within $\mathbf {CL}_{Q_s}$ Error Zone
\end{tablenotes}
\end{table*}

\begin{table*}[!htb]
	\scriptsize
	\renewcommand{\arraystretch}{1.2}
	\caption{Target Battery SOH Estimation Results }
	\vspace{-3mm}
	\label{table_4}
	\begin{center}
	\setlength{\tabcolsep}{4.4pt}
			\begin{tabular}{>{\centering\arraybackslash}m{0.10\textwidth}|>{\centering\arraybackslash}m{0.06\textwidth}>{\centering\arraybackslash}m{0.06\textwidth}|>{\centering\arraybackslash}m{0.06\textwidth}>{\centering\arraybackslash}m{0.06\textwidth}|>{\centering\arraybackslash}m{0.06\textwidth}p{0.06\textwidth}>{\centering\arraybackslash}m{0.06\textwidth}>{\centering\arraybackslash}m{0.06\textwidth}|>{\centering\arraybackslash}m{0.09\textwidth}>{\centering\arraybackslash}m{0.09\textwidth}}
				\hline
				\toprule  
 				\multicolumn{1}{c|}{\textbf{Source Battery}} &\multicolumn{2}{c|}{\textbf{Target Battery}} &\multicolumn{2}{c|}{\textbf{SELF-GRU}} &\multicolumn{2}{c}{\textbf{DCVA-GRU}} &\multicolumn{2}{c|}{\textbf{DCVA-GRU-TL}} &\multicolumn{2}{c}{\textbf{Accuracy Improvement By TL}} \\
 				\multicolumn{1}{c|}{\textbf{}} &\multicolumn{2}{c|}{\textbf{}} &\multicolumn{2}{c|}{\textbf{}}&\multicolumn{2}{c}{\textbf{(Proposed-w/o TL)}} &\multicolumn{2}{c|}{\textbf{(Proposed-w/ TL)}} &\multicolumn{2}{c}{\textbf{(DCVA-GRU-TL VS DCVA-GRU)}} \\
				\toprule 
				Channel&Channel&Similarity&MAE&RMSE&MAE&RMSE&MAE&RMSE&MAE ($\%$) &RMSE ($\%$)\\
				\toprule  
				\multirow{4}{2em}{\centering CH01}&CH22&$\mathbf {Yes}$&0.0520&0.0909&0.0061&0.0077&$\mathbf {0.0026}$&$\mathbf {0.0034}$&$\mathbf {58\%}$&$\mathbf {55\%}$\\
				&CH29&$\mathbf {Yes}$&0.0546&0.0811&0.0197&0.0201&$\mathbf {0.0037}$&$\mathbf {0.0047}$&$\mathbf {81\% }$&$\mathbf {77\%}$\\
				&CH32&No&0.0527&0.0868&0.0188&0.0192&0.0079&0.0147&58$\%$&23$\%$\\
				&CH33&No&0.0541&0.0941&0.0140&0.0164&0.0081&0.0149&42$\%$&9$\%$\\
				\toprule 
				\multirow{4}{2em}{\centering CH21}&CH26&$\mathbf {Yes}$&0.0557&0.0905&0.0037&0.0046&$\mathbf {0.0033}$&$\mathbf {0.0040}$&$\mathbf {11\%}$&$\mathbf {12\%}$\\
				&CH45&$\mathbf {Yes}$&0.0500&0.0754&0.0114&0.0150&$\mathbf {0.0053}$&$\mathbf {0.0067}$&$\mathbf {53\%}$&$\mathbf {56\%}$\\
				&CH15&No&0.0522&0.0882&0.0151&0.0162&0.0087&0.0110&43$\%$&32$\%$\\
				&CH19&No&0.0639&0.1078&0.0344&0.0390&0.0181&0.0234&47$\%$&40$\%$\\
				\toprule 
				\multirow{4}{2em}{\centering CH25}&CH28&$\mathbf {Yes}$&0.0515&0.0858&0.0186&0.0188&$\mathbf {0.0034}$&$\mathbf {0.0044}$&$\mathbf {82\%}$&$\mathbf {77\%}$\\
				&CH41&$\mathbf {Yes}$&0.0588&0.0949&0.0150&0.0159&$\mathbf {0.0031}$&$\mathbf {0.0039}$&$\mathbf {79\%}$&$\mathbf {76\%}$\\
				&CH06&No&0.0485&0.0864&0.0160&0.0180&0.0133&0.0154&17$\%$&15$\%$\\
				&CH47&No&0.0371&0.0604&0.0122&0.0149&0.0109&0.0146&10$\%$&2$\%$\\
				\toprule 
				\multirow{4}{2em}{\centering CH37}&CH15&$\mathbf {Yes}$&0.0521&0.0882&0.0180&0.0182&$\mathbf {0.0070}$&$\mathbf {0.0079}$&$\mathbf {61\%}$&$\mathbf {57\%}$\\
				&CH27&$\mathbf {Yes}$&0.0555&0.0917&0.0130&0.0133&$\mathbf {0.0034}$&$\mathbf {0.0043}$&$\mathbf {74\%}$&$\mathbf {67\%}$\\
				&CH02&No&0.0521&0.0906&0.0166&0.0202&0.0106&0.0166&36$\%$&18$\%$\\
				&CH03&No&0.0516&0.0906&0.0272&0.0303&0.0177&0.0231&35$\%$&24$\%$\\
				\hline
				\toprule
			\end{tabular}
	\end{center}
	\vspace{-2mm}
\vspace{-4mm}
\end{table*}

\begin{figure*}[htb]
\centering
\includegraphics[scale=0.55]{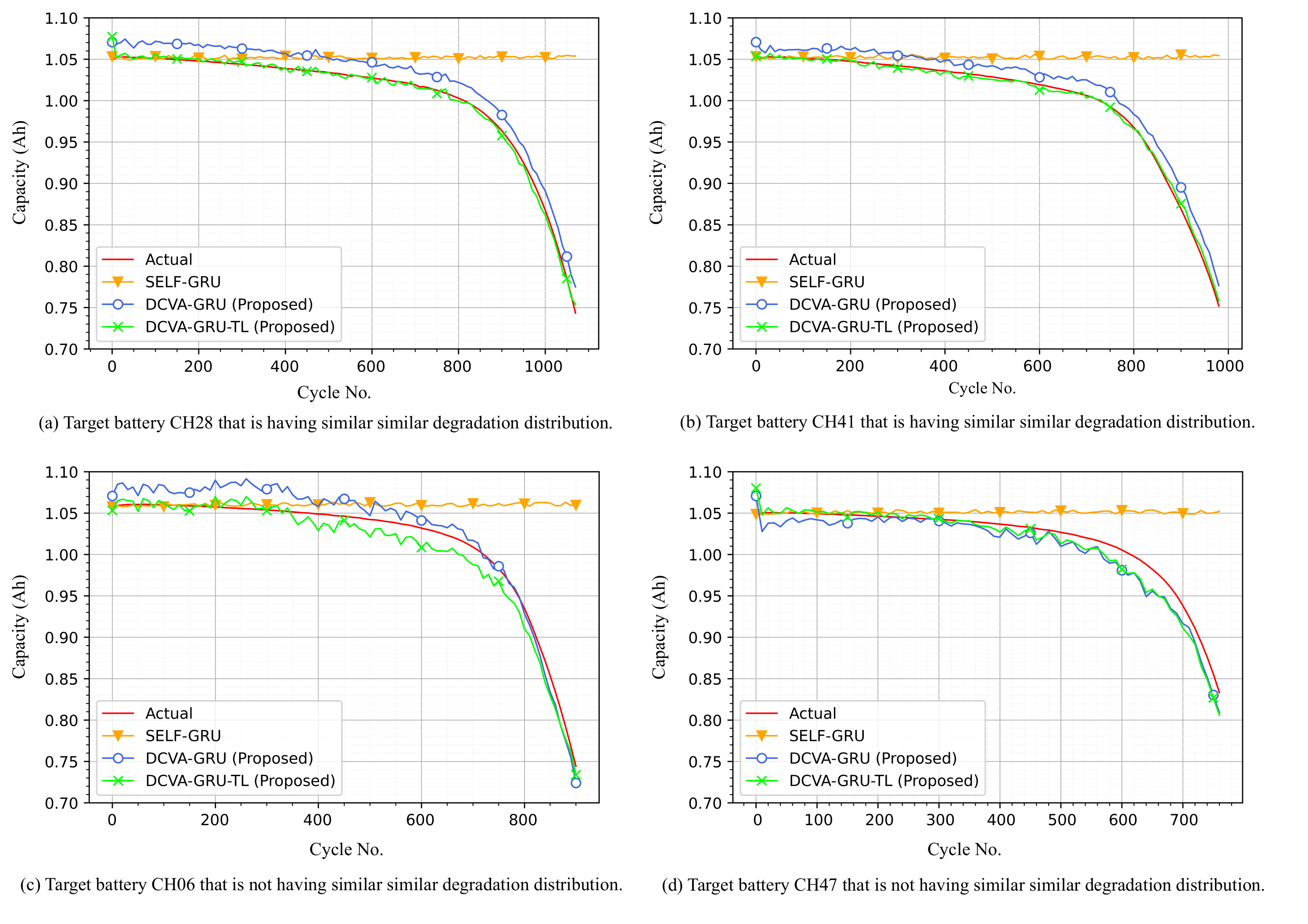}
\caption{Based on source battery CH25,  estimation result comparison between the proposed method and direct apply}
\label{MyFig9}
\end{figure*}

\subsubsection{Target SOH Estimation Model}
The proposed TL target battery SOH estimation method DCVA-GRU-TL is investigated here for those target batteries with a similar degradation distribution as the source battery.  The source battery SOH estimation model trained under CVs can be transferred to the target battery and compensated by the target battery residual model through its unique RVs using the target battery$'$s first 100 discharge voltage cycles only.  

"How to transfer" is finalized by combining both the source battery SOH estimation model and the target battery residual estimation model to perform target battery SOH estimation. 

The residual model is built by a GRU layer of 300 neurons followed by a dropout layer of 20$\%$, then another GRU layer of 500 neurons followed by a 20$\%$ dropout. Before the output layer, a normal layer with 100 neurons is inserted. Thirty epochs are run. The same optimizer Adam is used.  

Compared with the source battery SOH estimation model DCVA-GRU as described in Fig. 2(b), for those with similar degradation distribution,  DCVA-GRU-TL has achieved excellent estimation results to achieve RMSE going below 0.01 as shown in Table III.  While for those target batteries that are not similar to the source battery,  RMSE cannot go below 0.01 even after target battery residual model compensation. Using source battery CH25 for illustration, similar target battery CH28 MAE and RMSE improved by 82$\%$ and 77$\%$, hitting 0.0034 and 0.0044, while CH41 MAE and RMSE improved by 79$\%$ and 76$\%$, going as low as 0.0031 and 0.0039. For CH06, which is not similar, although 15$\%$ RMSE improvements are recorded, RMSE is still 0.0154. The same finding is for CH47,  where RMSE only improves by 2$\%$ to 0.0146.  Both RMSEs are significantly higher than 0.01. The model described in Fig. 2(a) was also put to the test. With only the first 100 discharge voltage cycles, it fails to predict the SOH with such a small training dataset that MAE is larger than 0.05 and RMSE is greater than 0.08. For better visualization, the detailed prediction trends of all methods are illustrated in Fig. 9.  Compensated by the target battery's unique feature,  the proposed method DCVA-GRU-TL yields the best estimation result of RMSE below 0.01 for the battery having a similar degradation distribution as the source battery.

\section{Conclusion}
This research utilizes temporal dynamic correlation during discharging and presents an innovative TL-based SOH estimation approach.  Dynamic time warping translates the target battery discharge voltage-based time series into a time-index-based time series based on the source battery's discharge voltage cycle at full capacity. It synchronizes discharge data while also extracting temporal information.  Canonical variate analysis underlines the temporal correlation within the cycles and separates them into common and domain-specific features.  The source SOH estimation model is built by the common feature and its ability to estimate the target battery SOH is assessed by statistic indicators.  It serves as a benchmark for the target batteries to choose the best source SOH estimation model for a better estimation result.  Based on the outcome of the assessment , target battery is able to select the best source SOH estimation model from source battery pool to optimize the estimation accuracy.  Then, using common feature canonical variates,  transfer learning functioned as a bridge to transfer the knowledge obtained by the source SOH estimate model, which was trained by data from the complete degradation process. Estimation is further enhanced and precise when it is combined with the target residual SOH estimation model, built by the target battery's unique feature. The proposed methods are validated by a public dataset and show effectiveness for SOH estimation by using as little as the first 100 discharge cycles regardless of charging policy.  

Considering the importance of temperature on the battery performance aging, future work is recommended to design a generic source SOH model that involves the influence of varying ambient temperatures.

\bibliographystyle{IEEEtran}
\bibliography{Reference.bib}

\begin{thebibliography}{10}
\providecommand{\url}[1]{#1}
\csname url@samestyle\endcsname
\providecommand{\newblock}{\relax}
\providecommand{\bibinfo}[2]{#2}
\providecommand{\BIBentrySTDinterwordspacing}{\spaceskip=0pt\relax}
\providecommand{\BIBentryALTinterwordstretchfactor}{4}
\providecommand{\BIBentryALTinterwordspacing}{\spaceskip=\fontdimen2\font plus
\BIBentryALTinterwordstretchfactor\fontdimen3\font minus
  \fontdimen4\font\relax}
\providecommand{\BIBforeignlanguage}[2]{{%
\expandafter\ifx\csname l@#1\endcsname\relax
\typeout{** WARNING: IEEEtran.bst: No hyphenation pattern has been}%
\typeout{** loaded for the language `#1'. Using the pattern for}%
\typeout{** the default language instead.}%
\else
\language=\csname l@#1\endcsname
\fi
#2}}
\providecommand{\BIBdecl}{\relax}
\BIBdecl

\bibitem{58}
X.~Hu, Y.~Che, X.~Lin, and Z.~Deng, ``Health prognosis for electric vehicle
  battery packs: A data-driven approach,'' \emph{IEEE/ASME Transactions on
  Mechatronics}, vol.~25, no.~6, pp. 2622--2632, 2020.

\bibitem{61}
H.~Wang, Y.~Huang, and A.~Khajepour, ``Cyber-physical control for energy
  management of off-road vehicles with hybrid energy storage systems,''
  \emph{IEEE/ASME Transactions on Mechatronics}, vol.~23, no.~6, pp.
  2609--2618, 2018.

\bibitem{57}
Y.~Gao, X.~Zhang, C.~Zhu, and B.~Guo, ``Global parameter sensitivity analysis
  of electrochemical model for lithium-ion batteries considering aging,''
  \emph{IEEE/ASME Transactions on Mechatronics}, vol.~26, no.~3, pp.
  1283--1294, 2021.

\bibitem{33}
Y.~Qin, C.~Yuen, and S.~Adams, ``Invariant learning based multi-stage
  identification for lithium-ion battery performance degradation,'' \emph{The
  46th Annual Conference of the IEEE Industrial Electronics Society}, pp.
  1849--1854, 2020.

\bibitem{53}
G.~Kang, L.~Wu, Y.~Guan, and Z.~Peng, ``A virtual sample generation method
  based on differential evolution algorithm for overall trend of small sample
  data: Used for lithium-ion battery capacity degradation data,'' \emph{IEEE
  Access}, vol.~7, pp. 123\,255--123\,267, 2019.

\bibitem{56}
A.~Khalil, K.~F. Aljanaideh, and M.~A. Janaideh, ``Output-only measurements for
  fault detection of multi-actuator systems in motion control applications,''
  \emph{IEEE Sensors Journal}, vol.~22, no.~5, pp. 4164--4174, 2022.

\bibitem{34}
Y.~Qin, W.~Li, C.~Yuen, W.~Tushar, and T.~Saha, ``Iiot-enabled health
  monitoring for integrated heat pump system using mixture slow feature
  analysis,'' \emph{IEEE Transactions on Industrial Informatics}, pp. 1--1,
  2021.

\bibitem{55}
L.~A.~Q. Cordovil, P.~H.~S. Coutinho, I.~V. de~Bessa, M.~F. S.~V. D'Angelo, and
  R.~M. Palhares, ``Uncertain data modeling based on evolving ellipsoidal fuzzy
  information granules,'' \emph{IEEE Transactions on Fuzzy Systems}, vol.~28,
  no.~10, pp. 2427--2436, 2020.

\bibitem{51}
R.~Rahimilarki, Z.~Gao, A.~Zhang, and R.~Binns, ``Robust neural network fault
  estimation approach for nonlinear dynamic systems with applications to wind
  turbine systems,'' \emph{IEEE Transactions on Industrial Informatics},
  vol.~15, no.~12, pp. 6302--6312, 2019.

\bibitem{25}
X.~Feng, C.~Weng, X.~He, X.~Han, L.~Lu, D.~Ren, and M.~Ouyang, ``Online
  state-of-health estimation for li-ion battery using partial charging segment
  based on support vector machine,'' \emph{IEEE Transactions on Vehicular
  Technology}, vol.~68, no.~9, pp. 8583--8592, 2019.

\bibitem{60}
Z.~Deng, X.~Hu, X.~Lin, L.~Xu, Y.~Che, and L.~Hu, ``General discharge voltage
  information enabled health evaluation for lithium-ion batteries,''
  \emph{IEEE/ASME Transactions on Mechatronics}, vol.~26, no.~3, pp.
  1295--1306, 2021.

\bibitem{48}
M.~Raman, V.~Champa, and V.~Prema, ``State of health estimation of lithium ion
  batteries using recurrent neural network and its variants,'' \emph{2021 IEEE
  International Conference on Electronics, Computing and Communication
  Technologies (CONECCT)}, pp. 1--6, 2021.

\bibitem{3}
S.~Hochreiter and J.~Schmidhuber, ``Long short-term memory,'' \emph{Neural
  computation}, vol.~9, no.~8, pp. 1735--1780, 1997.

\bibitem{4}
K.~Cho, B.~Van~Merri{\"e}nboer, C.~Gulcehre, D.~Bahdanau, F.~Bougares,
  H.~Schwenk, and Y.~Bengio, ``Learning phrase representations using rnn
  encoder-decoder for statistical machine translation,'' \emph{arXiv preprint
  arXiv:1406.1078}, 2014.

\bibitem{68}
Y.~Qin, C.~Yuen, Y.~Shao, B.~Qin, and X.~Li, ``Slow-varying dynamics-assisted
  temporal capsule network for machinery remaining useful life estimation,''
  \emph{IEEE Transactions on Cybernetics}, pp. 1--15, 2022, doi:
  10.1109/TCYB.2022.3164683.

\bibitem{15}
L.~Jayasinghe, T.~Samarasinghe, C.~Yuen, J.~C. Ni~Low, and S.~Sam~Ge,
  ``Temporal convolutional memory networks for remaining useful life estimation
  of industrial machinery,'' in \emph{2019 IEEE International Conference on
  Industrial Technology (ICIT)}, 2019, pp. 915--920.

\bibitem{16}
Y.~Zhang, R.~Xiong, H.~He, and M.~G. Pecht, ``Long short-term memory recurrent
  neural network for remaining useful life prediction of lithium-ion
  batteries,'' \emph{IEEE Transactions on Vehicular Technology}, vol.~67,
  no.~7, pp. 5695--5705, 2018.

\bibitem{14}
S.~Cui and I.~Joe, ``A dynamic spatial-temporal attention-based gru model with
  healthy features for state-of-health estimation of lithium-ion batteries,''
  \emph{IEEE Access}, vol.~9, pp. 27\,374--27\,388, 2021.

\bibitem{47}
T.~Han, Y.-F. Li, and M.~Qian, ``A hybrid generalization network for
  intelligent fault diagnosis of rotating machinery under unseen working
  conditions,'' \emph{IEEE Transactions on Instrumentation and Measurement},
  vol.~70, pp. 1--11, 2021.

\bibitem{52}
F.~Abate, V.~K.~L. Huang, G.~Monte, V.~Paciello, and A.~Pietrosanto, ``A
  comparison between sensor signal preprocessing techniques,'' \emph{IEEE
  Sensors Journal}, vol.~15, no.~5, pp. 2479--2487, 2015.

\bibitem{59}
Y.~Ding, P.~Ding, X.~Zhao, Y.~Cao, and M.~Jia, ``Transfer learning for
  remaining useful life prediction across operating conditions based on
  multisource domain adaptation,'' \emph{IEEE/ASME Transactions on
  Mechatronics}, pp. 1--10, 2022.

\bibitem{7}
X.~Shu, J.~Shen, G.~Li, Y.~Zhang, Z.~Chen, and Y.~Liu, ``A flexible
  state-of-health prediction scheme for lithium-ion battery packs with long
  short-term memory network and transfer learning,'' \emph{IEEE Transactions on
  Transportation Electrification}, vol.~7, no.~4, pp. 2238--2248, 2021.

\bibitem{8}
Y.~Tan and G.~Zhao, ``Transfer learning with long short-term memory network for
  state-of-health prediction of lithium-ion batteries,'' \emph{IEEE
  Transactions on Industrial Electronics}, vol.~67, no.~10, pp. 8723--8731,
  2020.

\bibitem{9}
Y.~Che, Z.~Deng, X.~Lin, L.~Hu, and X.~Hu, ``Predictive battery health
  management with transfer learning and online model correction,'' \emph{IEEE
  Transactions on Vehicular Technology}, vol.~70, no.~2, pp. 1269--1277, 2021.

\bibitem{64}
S.~Kim, Y.~Y. Choi, K.~J. Kim, and J.-I. Choi, ``Forecasting state-of-health of
  lithium-ion batteries using variational long short-term memory with transfer
  learning,'' \emph{Journal of Energy Storage}, vol.~41, p. 102893, 2021.

\bibitem{65}
G.~Ma, S.~Xu, T.~Yang, Z.~Du, L.~Zhu, H.~Ding, and Y.~Yuan, ``A transfer
  learning-based method for personalized state of health estimation of
  lithium-ion batteries,'' \emph{IEEE Transactions on Neural Networks and
  Learning Systems}, pp. 1--11, 2022, doi: 10.1109/TNNLS.2022.3176925.

\bibitem{17}
Y.~Choi, S.~Ryu, K.~Park, and H.~Kim, ``Machine learning-based lithium-ion
  battery capacity estimation exploiting multi-channel charging profiles,''
  \emph{IEEE Access}, vol.~7, pp. 75\,143--75\,152, 2019.

\bibitem{19}
Z.~Chen, M.~Wu, R.~Zhao, F.~Guretno, R.~Yan, and X.~Li, ``Machine remaining
  useful life prediction via an attention-based deep learning approach,''
  \emph{IEEE Transactions on Industrial Electronics}, vol.~68, no.~3, pp.
  2521--2531, 2021.

\bibitem{20}
Y.~Wu, M.~Yuan, S.~Dong, L.~Lin, and Y.~Liu, ``Remaining useful life estimation
  of engineered systems using vanilla lstm neural networks,''
  \emph{Neurocomputing}, vol. 275, pp. 167--179, 2018.

\bibitem{44}
H.~Hotelling, ``Relations between two sets of variates,'' in
  \emph{Breakthroughs in statistics}.\hskip 1em plus 0.5em minus 0.4em\relax
  Springer, 1992, pp. 162--190.

\bibitem{63}
C.~Zhang, Y.~Zhang, and Y.~Li, ``A novel battery state-of-health estimation
  method for hybrid electric vehicles,'' \emph{IEEE/ASME Transactions on
  Mechatronics}, vol.~20, no.~5, pp. 2604--2612, 2015.

\bibitem{5}
X.~Zhang, Y.~Qin, C.~Yuen, L.~Jayasinghe, and X.~Liu, ``Time-series
  regeneration with convolutional recurrent generative adversarial network for
  remaining useful life estimation,'' \emph{IEEE Transactions on Industrial
  Informatics}, vol.~17, no.~10, pp. 6820--6831, 2021.

\bibitem{21}
P.~M. Attia, A.~Grover, N.~Jin, K.~A. Severson, T.~M. Markov, Y.-H. Liao, M.~H.
  Chen, B.~Cheong, N.~Perkins, Z.~Yang \emph{et~al.}, ``Closed-loop
  optimization of fast-charging protocols for batteries with machine
  learning,'' \emph{Nature}, vol. 578, no. 7795, pp. 397--402, 2020.

\bibitem{12}
K.~Liu, Y.~Li, X.~Hu, M.~Lucu, and W.~D. Widanage, ``Gaussian process
  regression with automatic relevance determination kernel for calendar aging
  prediction of lithium-ion batteries,'' \emph{IEEE Transactions on Industrial
  Informatics}, vol.~16, no.~6, pp. 3767--3777, 2020.

\bibitem{38}
K.~Weiss, T.~M. Khoshgoftaar, and D.~Wang, ``A survey of transfer learning,''
  \emph{Journal of Big data}, vol.~3, no.~1, pp. 1--40, 2016.

\bibitem{50}
Z.~Gao, C.~Cecati, and S.~X. Ding, ``A survey of fault diagnosis and
  fault-tolerant techniques---part i: Fault diagnosis with model-based and
  signal-based approaches,'' \emph{IEEE Transactions on Industrial
  Electronics}, vol.~62, no.~6, pp. 3757--3767, 2015.

\bibitem{1}
K.~Q. Zhou, Y.~Qin, B.~P.~L. Lau, C.~Yuen, and S.~Adams, ``Lithium-ion battery
  state of health estimation based on cycle synchronization using dynamic time
  warping,'' in \emph{47th Annual Conference of the IEEE Industrial Electronics
  Society (IES)}, 2021.

\bibitem{2}
A.~Kassidas, J.~F. MacGregor, and P.~A. Taylor, ``Synchronization of batch
  trajectories using dynamic time warping,'' \emph{AIChE Journal}, vol.~44,
  no.~4, pp. 864--875, 1998.

\bibitem{28}
B.~Jiang, X.~Zhu, D.~Huang, and R.~D. Braatz, ``Canonical variate
  analysis-based monitoring of process correlation structure using causal
  feature representation,'' \emph{Journal of Process Control}, vol.~32, pp.
  109--116, 2015.

\bibitem{45}
M.~Beaghen, \emph{Canonical variate analysis and related methods with
  longitudinal data}.\hskip 1em plus 0.5em minus 0.4em\relax Virginia
  Polytechnic Institute and State University, 1997.

\bibitem{62}
Y.~Qin, S.~Adams, and C.~Yuen, ``Transfer learning-based state of charge
  estimation for lithium-ion battery at varying ambient temperatures,''
  \emph{IEEE Transactions on Industrial Informatics}, vol.~17, no.~11, pp.
  7304--7315, 2021.

\bibitem{41}
C.~Ruiz-C{\'a}rcel, Y.~Cao, D.~Mba, L.~Lao, and R.~Samuel, ``Statistical
  process monitoring of a multiphase flow facility,'' \emph{Control Engineering
  Practice}, vol.~42, pp. 74--88, 2015.

\bibitem{46}
K.~E.~S. Pilario and Y.~Cao, ``Canonical variate dissimilarity analysis for
  process incipient fault detection,'' \emph{IEEE Transactions on Industrial
  Informatics}, vol.~14, no.~12, pp. 5308--5315, 2018.

\bibitem{39}
P.-E.~P. Odiowei and Y.~Cao, ``Nonlinear dynamic process monitoring using
  canonical variate analysis and kernel density estimations,'' \emph{IEEE
  Transactions on Industrial Informatics}, vol.~6, no.~1, pp. 36--45, 2010.

\bibitem{67}
K.~He and J.~Sun, ``Convolutional neural networks at constrained time cost,''
  in \emph{Proceedings of the IEEE conference on computer vision and pattern
  recognition}, 2015, pp. 5353--5360.

\bibitem{66}
S.~Greenbank and D.~Howey, ``Automated feature extraction and selection for
  data-driven models of rapid battery capacity fade and end of life,''
  \emph{IEEE Transactions on Industrial Informatics}, vol.~18, no.~15, pp.
  2965--2973, 2022.

\end{thebibliography}

\vspace{-11mm}

\begin{IEEEbiography}[{\includegraphics[width=1in,height=1.25in,clip,keepaspectratio]{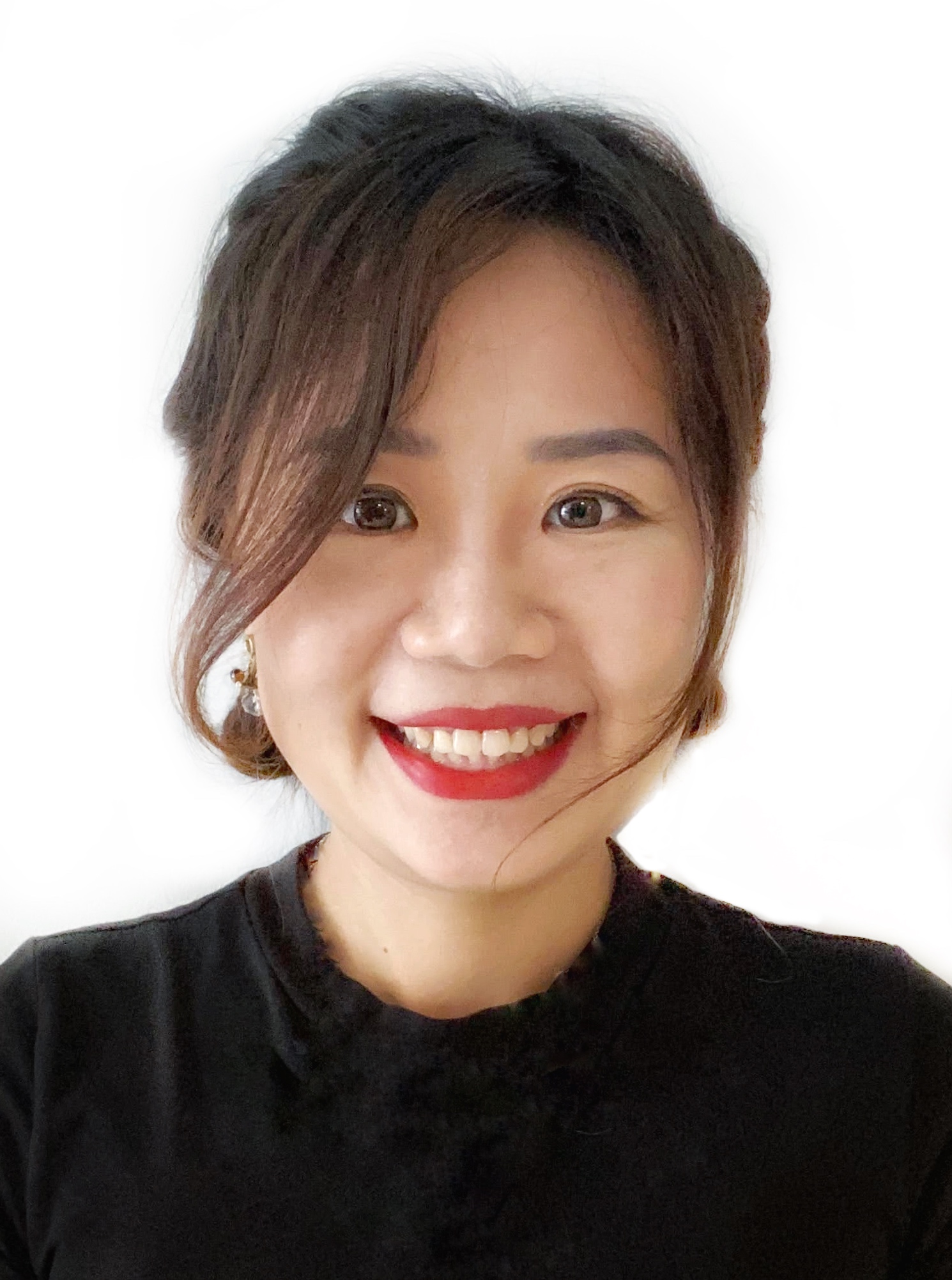}}]{Kate Qi Zhou}
received the B.S. degree in control science and engineering from Zhejiang University, Hangzhou, China, in 2002, and the M.S. degree in Innovation in Manufacturing Systems and Technology (IMST) under Singapore-MIT Alliance (SMA) program from Nanyang Technological University, Singapore, in 2003.  She is now pursuing her Ph.D. at Singapore University of Technology and Design, Singapore. Her research interests include time series analysis, data-driven process monitoring, Industrial Internet of Things,  digital twin, and state of health estimation for electronic vehicle batteries.
\end{IEEEbiography}

\vspace{-11mm}

\begin{IEEEbiography}[{\includegraphics[width=1in,height=1.25in,clip,keepaspectratio]{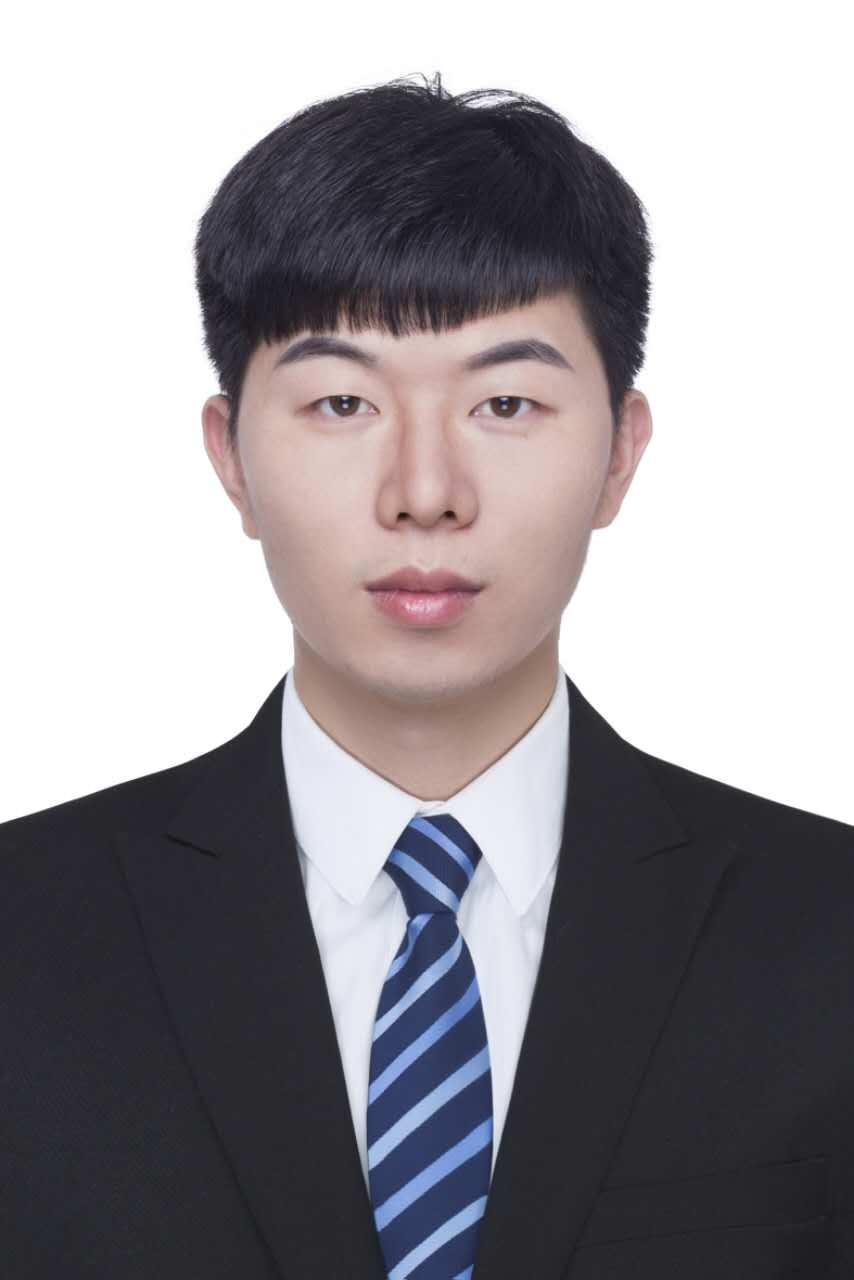}}]{Yan Qin}
received the B.S. degree in electronic information engineering from Information Engineering University, Zhengzhou, China, in 2011, the M.S. degree in control theory and control engineering from Northeastern University, Shenyang, China, in 2013, and the Ph.D. degree in control science and engineering from Zhejiang University, Hangzhou, China, in 2018. He is currently a Post-Doctoral Research Fellow at the Singapore University of Technology and Design. His research interests include data-driven process monitoring, soft sensor modeling, and remaining useful life estimation for industrial processes and essential assets.
\end{IEEEbiography}

\vspace{-11mm}

\begin{IEEEbiography}[{\includegraphics[width=1in,height=1.25in,clip,keepaspectratio]{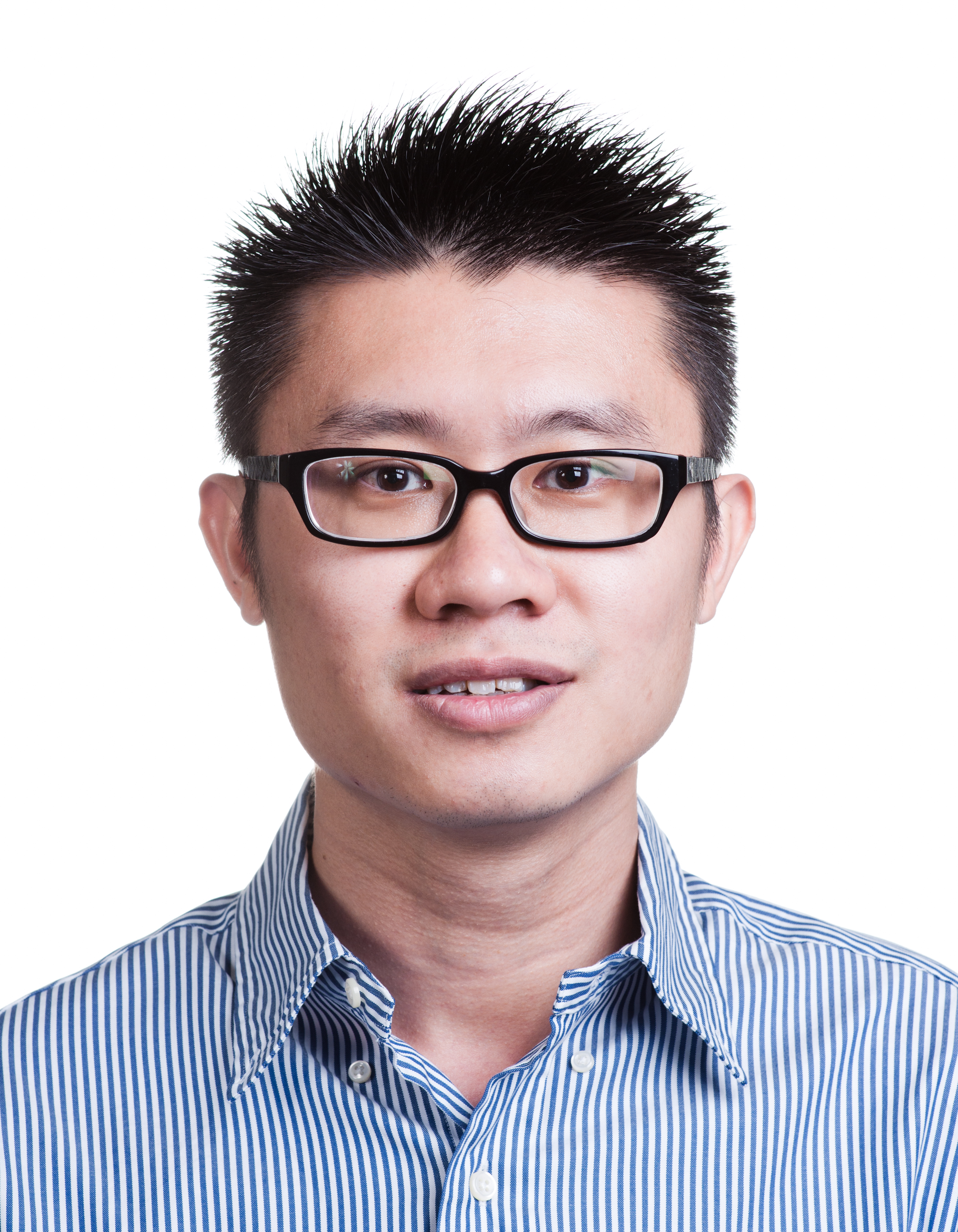}}]{Chau Yuen}
received the B.Eng. and Ph.D. degrees in information and communication from Nanyang Technological University, Singapore, in 2000 and 2004, respectively. He was a Post-Doctoral Fellow with Lucent Technologies Bell Labs, Murray Hill, in 2005, and a Visiting Assistant Professor with The Hong Kong Polytechnic University in 2008. From 2006 to 2010, he was with the Institute for Infocomm Research, Singapore. Since 2010, he has been with the Singapore University of Technology and Design.

Dr. Yuen was a recipient of the Lee Kuan Yew Gold Medal, the Institution of Electrical Engineers Book Prize, the Institute of Engineering of Singapore Gold Medal, the Merck Sharp and Dohme Gold Medal, and twice a recipient of the Hewlett Packard Prize. He received the IEEE Asia Pacific Outstanding Young Researcher Award in 2012 and IEEE VTS Singapore Chapter Outstanding Service Award on 2019. He serves as an Editor for the IEEE Transactions on Communications, and the IEEE Transactions on Vehicular Technology, where he was awarded as the Top Associate Editor from 2009 to 2015. He served as the Guest Editor for several special issues, including IEEE Journal on Selected Areas in Communications, IEEE Communications Magazine, IEEE Transaction on Cognitive Communications and Networking. He is a Distinguished Lecturer of IEEE Vehicular Technology Society.
\end{IEEEbiography}

\vspace{12pt}
\end{document}